\documentclass[11pt]{article}

\usepackage[top=25mm, bottom=25mm, left=25mm, right=25mm]{geometry}
\usepackage{authblk}
\usepackage{graphicx}
\usepackage[export]{adjustbox}
\usepackage[utf8]{inputenc}
\usepackage{xcolor}
\usepackage{subcaption}
\usepackage{booktabs}
\usepackage{makecell}
\usepackage{caption}
\usepackage{placeins}
\usepackage{amssymb}
\usepackage{amsmath}
\usepackage{float}
\usepackage{mdframed}
\usepackage[hidelinks]{hyperref}

\newcommand{\dataclinic}{\textit{Data Clinic}}

\newcommand{\lsexplorer}{\textit{Latent Space Explorer}}
\newcommand{\tiled}{\textit{Tiled}}

\newcommand{\mlexcompute}{\textit{MLExCompute}}
\newcommand{\filemanager}{\textit{File Manager}}
\newcommand{\pytorch}{\textit{PyTorch}}

\newcommand{\mlex}{\textit{MLExchange}}
\newcommand{\prefect}{\textit{Prefect}}
\newcommand{\mlflow}{\textit{MLflow}}
\newcommand{\bluesky}{\textit{Bluesky}}
\newcommand{\arroyo}{\textit{ArroyoPy}}

\begin{document}

\title{Unlocking Latent Dimensions: Exploring Representations of
  Large-Scale X-ray Scattering Data using Variational Autoencoders}

\author[1]{Monika Choudhary}
\author[1]{Xiaoya Chong}
\author[1]{Runbo Jiang}
\author[1]{Wiebke Koepp}
\author[2,3,4]{Petrus H. Zwart}
\author[1]{Damon English}
\author[1,5]{Gregory M. Su}
\author[1]{Eric Schaible}
\author[1]{Chenhui Zhu}
\author[7]{Mostafa Nassr}
\author[7]{Noah P. Wamble}
\author[1,5,8]{Kelvin Kam-Yun Li}
\author[1]{Jonathan M. Chan}
\author[1,7]{Jose Carlos Diaz}
\author[6]{Cameron McKay}
\author[6]{Lynn Katz}
\author[7]{Benny Freeman}
\author[9,10]{Guillaume Freychet}
\author[9]{Yevgen Matviychuk}
\author[9]{Eliot Gann}
\author[9]{Daniel B. Allan}
\author[1,11]{Benedikt Sochor}
\author[11]{Frank Schluenzen}
\author[11,12]{Stephan V. Roth}
\author[1,13]{Ethan J. Crumlin}
\author[1]{Dylan McReynolds}
\author[1]{Tanny Chavez\thanks{Corresponding author: \texttt{tanchavez@lbl.gov}}}
\author[1]{Alexander Hexemer\thanks{Corresponding author: \texttt{ahexemer@lbl.gov}}}

\affil[1]{Advanced Light Source, Lawrence Berkeley National Laboratory, Berkeley, CA, USA}
\affil[2]{Center for Advanced Mathematics for Energy Research Applications, Lawrence Berkeley National Laboratory, Berkeley, CA, USA}
\affil[3]{Molecular Biophysics \& Integrated Bioimaging Division, Lawrence Berkeley National Laboratory, Berkeley, CA, USA}
\affil[4]{Berkeley Synchrotron Infrared Structural Biology Program, Lawrence Berkeley National Laboratory, Berkeley, CA, USA}
\affil[5]{Materials Sciences Division, Lawrence Berkeley National Laboratory, Berkeley, CA, USA}
\affil[6]{Maseeh Department of Civil, Architectural, and Environmental Engineering, University of Texas, Austin, TX, USA}
\affil[7]{McKetta Department of Chemical Engineering, University of Texas, Austin, TX, USA}
\affil[8]{Department of Chemistry, University of California, Berkeley, CA, USA}
\affil[9]{National Synchrotron Light Source II, Brookhaven National Laboratory, Upton, NY, USA}
\affil[10]{University Grenoble Alpes, CEA, Leti, F-38000 Grenoble, France}
\affil[11]{Deutsches Elektronen-Synchrotron DESY, Notkestr.~85, 22607 Hamburg, Germany}
\affil[12]{Department of Fibre and Polymer Technology, Royal Institute of Technology KTH, Teknikringen 56, Stockholm, Sweden}
\affil[13]{Chemical Sciences Division, Lawrence Berkeley National Laboratory, Berkeley, CA, USA}

\date{}

\maketitle

\begin{abstract}
\noindent
Scientific user facilities generate X-ray scattering data faster than traditional workflows can process them. We address this challenge across two settings, offline dataset exploration and live on-the-fly analysis. We train a domain-specific attention-based Convolutional Variational Autoencoder (C-VAE) on 1.5~million X-ray scattering images to learn low-dimensional representations capturing structural variation across diverse experimental conditions. The learned latent space reveals well-organized clusters and smooth trajectories reflecting experimental progression. It further supports controlled synthetic scattering image generation across diverse structural states. When deployed without retraining, the model organizes time-resolved film formation experiments at two synchrotron facilities into interpretable latent structures. Benchmarking against DINOv3 (ViT-7B), a general-purpose vision foundation model, demonstrates that domain-specific training yields more interpretable latent
organization for scattering data. Both workflows are integrated
within \lsexplorer, a component of the \mlex\ platform, supporting interactive structural exploration across archived datasets and live experiments.
\end{abstract}


\section{Introduction}
\label{sec:intro}

The rapid growth of data volumes at modern experimental facilities has fundamentally changed how
experimental data is collected and analyzed at synchrotron facilities. Detectors at Scientific User
Facilities (SUFs) generate millions of high-resolution images under diverse experimental conditions,
resulting in heterogeneous, high-dimensional datasets. Traditional workflows based on manual inspection
or offline batch processing are increasingly insufficient to keep pace with modern data acquisition
rates. As experimental workflows move toward autonomous and adaptive data collection, there is a growing
need for computational models and software infrastructure capable of extracting interpretable structure
from large scientific datasets in real time~\cite{barbour2022advancing,parkinson2024ai,noack2019kriging}.
Recent work has shown that machine learning (ML) can help analyze complex X-ray data and reveal
underlying dynamics~\cite{horwath2024ai}, including classification of large diffraction
datasets~\cite{salgado2023automated}, physics-aware real-time analysis of nanodiffraction
patterns~\cite{luo2025donut}, and closed-loop feedback control at synchrotron
beamlines~\cite{pithan2023closing}. In particular, representation learning techniques based on deep
neural networks enable the discovery of meaningful structure in high-dimensional scientific data without
requiring domain experts to manually define or select the specific data features to be analyzed. For
imaging data, architectures such as Convolutional Neural Networks
(CNNs)~\cite{krizhevsky2012imagenet} effectively capture local spatial correlations, while attention
mechanisms~\cite{vaswani2017attention} extend this capability by modeling long-range dependencies and
hierarchical structure~\cite{dosovitskiy2020image,caron2021emerging}. Building on these advances,
modern approaches aim to encode complex data into compact, low-dimensional latent representations that
capture the dominant variations within a dataset.

Such latent representations provide a natural framework for organizing, visualizing, and exploring large
collections of scattering data~\cite{kingma2013auto,mcinnes2018umap}. Variational Autoencoders
(VAEs)~\cite{kingma2013auto} offer a probabilistic approach to learning these structured latent spaces
and have been successfully applied to a wide range of scientific imaging problems, including materials
characterization and scattering
experiments~\cite{cohn2021unsupervised,lee2020deep,kim2021exploration,huang2020interactive,valleti2024physics,kalinin2021exploring,calvat2025learning}
and manifold-aware synthetic data generation~\cite{chadebec2021data}. In parallel, self-supervised
vision transformer (ViT) models such as Distillation with No Labels (DINO) have demonstrated strong
performance in learning transferable image
representations~\cite{dosovitskiy2020image,oquab2023dinov2}. Together, these learned representations
enable a broad range of downstream tasks, including structural phase
tracking~\cite{sutherland2025autosas}, unsupervised anomaly
detection~\cite{strohmann2023can}, segmentation~\cite{ren2017fly,zhang2024towards}, and interactive
region-of-interest selection~\cite{seifi2025featureforest}. The generative capability of learned latent
spaces further enables synthetic scattering image generation, offering a route to augmenting
underrepresented structural states in experimental
datasets~\cite{zhao2024generating,chadebec2021data,kim2021exploration}. These representations are
commonly visualized using dimensionality reduction techniques such as Principal Component Analysis
(PCA)~\cite{mackiewicz1993principal}, Uniform Manifold Approximation and Projection
(UMAP)~\cite{mcinnes2018umap}, and t-distributed stochastic neighbor embedding
(t-SNE)~\cite{van2008visualizing}, enabling visualization of global relationships between observations.
Clustering methods such as HDBSCAN~\cite{campello2013density} further allow identification of
structurally similar patterns within large datasets. Interactive tools for navigating these learned
representations have been developed across several domains, including biomedical
imaging~\cite{kwon2023latent} and molecular discovery~\cite{zhang2024chemnav}, demonstrating the value
of visual analytics for high-dimensional scientific data.

Despite these advances, extracting interpretable structure from large-scale scattering datasets remains
challenging in two distinct settings. In \textit{post-experiment analysis}, where all data are available
offline, latent representations enable interactive exploration of large experimental archives. The
\mlex\ platform~\cite{zhao2022mlexchange} is a web-based environment for exchangeable machine learning
workflows at scientific user facilities, and its
\lsexplorer\footnote{\url{https://github.com/mlexchange/mlex_latent_explorer}} component provides
interactive dimensionality reduction, clustering, and visualization of learned embeddings for offline
dataset exploration~\cite{chavez2025machine}. In \textit{on-the-fly analysis}, data arrive
continuously during an active experiment, requiring a pre-trained model to be deployed before data
acquisition begins. Recent efforts have demonstrated ML-guided on-the-fly analysis in related scattering
settings, including autonomous phase identification in X-ray
diffraction~\cite{szymanski2023adaptively} and real-time structural tracking during thin film
crystallization~\cite{starostin2022tracking}. In this setting, a key question arises: is a
general-purpose vision model sufficient, or does the domain-specific character of X-ray scattering data
require a model trained specifically on scattering images. Domain-specific models have shown clear
advantages for scattering data; for example, a dedicated denoising model for SAXS/WAXD images
outperforms general-purpose approaches by capturing scattering-specific textural
features~\cite{zhou2023machine}. General-purpose self-supervised models learn powerful representations from large and diverse natural image collections, demonstrating remarkable capabilities across a wide range of vision tasks. However, whether such representations capture the structural variations most relevant for scientific imaging domains such as X-ray scattering remains an open question. To investigate this, we benchmark against DINOv3~\cite{simeoni2025dinov3}, a large vision foundation model extending the DINO self-supervised training framework~\cite{caron2021emerging,oquab2023dinov2} to ViT-7B scale with a Gram anchoring objective for improved feature consistency. DINOv3 was selected as the general-purpose baseline as its self-supervised ViT architecture most closely mirrors our attention-based C-VAE, ensuring performance differences reflect training data rather than architectural choices. We note that our evaluation uses global image embeddings, the regime where DINOv3 performs strongest, and any observed advantage for the domain-specific C-VAE should therefore be interpreted as a preliminary finding. In this work, we use the DINOv3 ViT-7B variant (approximately 7 billion parameters).

\begin{figure}[!htbp]
\centering
\fbox{%
  \includegraphics[width=\textwidth]{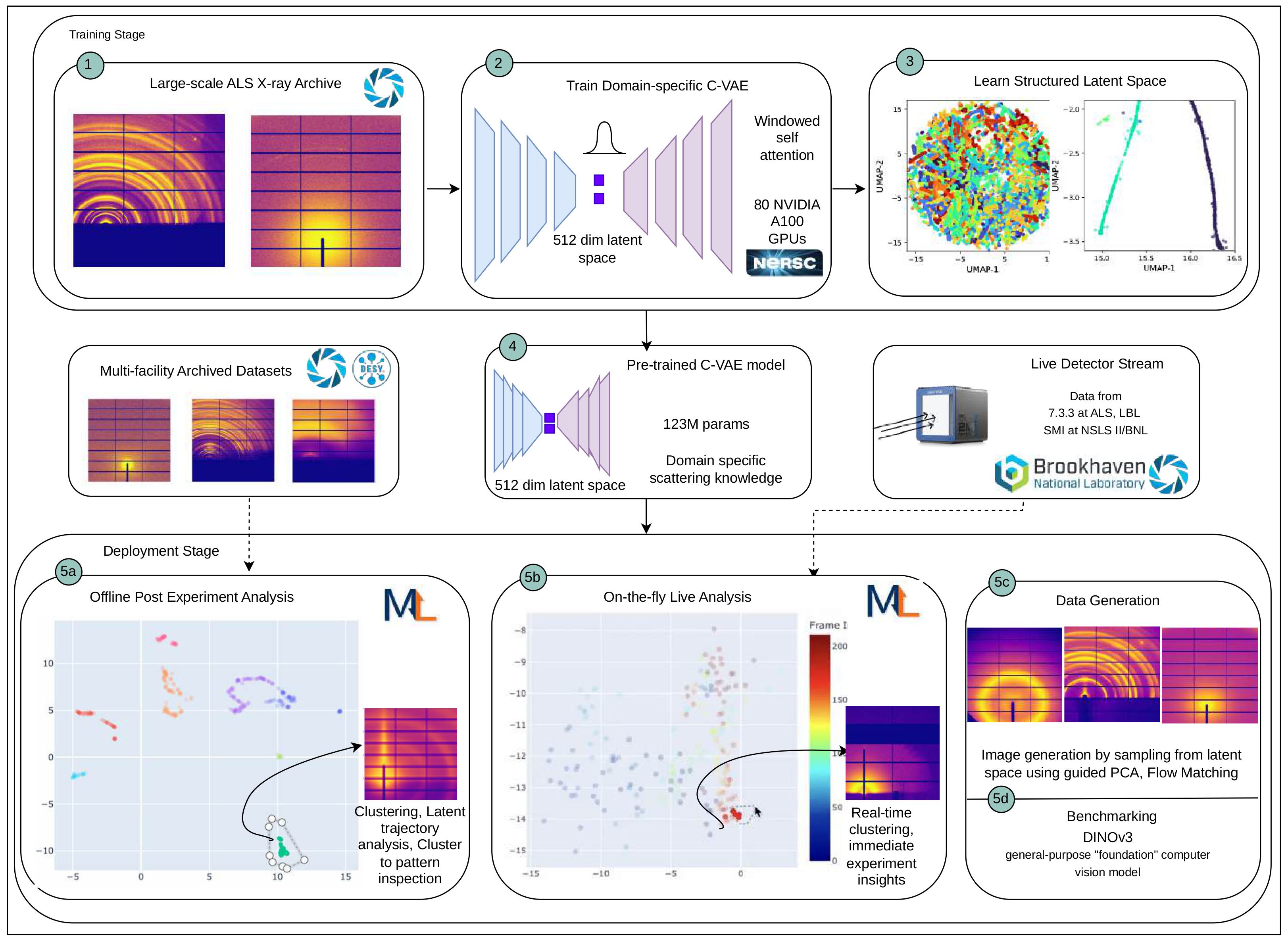}%
}
\caption{\textbf{C-VAE scattering data analysis pipeline.}
\textbf{(1)} A large-scale archive of 1.5 million X-ray scattering
images collected at the Advanced Light Source (ALS) serves as the
training dataset. \textbf{(2)} A domain-specific attention-based
convolutional variational autoencoder (C-VAE) with windowed
self-attention is trained on this archive using 80 NVIDIA A100 GPUs
at NERSC Perlmutter, learning a 512-dimensional latent representation
of the scattering data. \textbf{(3)} The learned latent space organizes
scattering patterns into structured clusters with smooth trajectories
reflecting experimental progression, visualized here via UMAP
projection. \textbf{(4)} The pre-trained C-VAE model (123M parameters,
512-dimensional latent space) is deployed without retraining across
downstream applications. \textbf{(5a)} Offline post-experiment analysis:
archived datasets from multiple facilities are interactively explored
through the Latent Space Explorer interface, supporting clustering,
latent trajectory analysis, and cluster-to-pattern inspection.
\textbf{(5b)} On-the-fly live analysis: detector images streamed during
active experiments at ALS beamline 7.3.3 and the NSLS-II SMI beamline
are embedded in real time, enabling immediate structural insights and
trajectory monitoring. \textbf{(5c)} Synthetic scattering image
generation via UMAP-guided PCA sampling and conditional flow matching.
\textbf{(5d)} Benchmarking against DINOv3, a general-purpose vision
foundation model, to evaluate the benefit of domain-specific training
for scattering data analysis.}
\label{fig:overview}
\end{figure}

In this work, we address both settings by training a domain-specific attention-based convolutional
variational autoencoder (C-VAE) on 1.5~million historical X-ray scattering images collected at the
Advanced Light Source. We first demonstrate that the C-VAE learns a well-organized latent space from
this historical dataset, in which clusters correspond to distinct scattering regimes and trajectories
reflect experimental progression. This pre-trained model is then deployed for on-the-fly analysis of
previously unseen experiments at two synchrotron facilities. To directly evaluate the benefit of
domain-specific training, we benchmark the C-VAE against DINOv3 (ViT-7B)~\cite{simeoni2025dinov3}, the most recent iteration of the DINO family of self-supervised vision transformers, succeeding DINOv2~\cite{oquab2023dinov2}. We use the same on-the-fly data to test whether a model trained on scattering images captures more interpretable latent structure than a large general-purpose vision model. Both workflows are integrated within the \lsexplorer\ component of the \mlex\ platform, supporting interactive exploration during both
offline post-experiment analysis and live experimental sessions.

This work makes the following contributions:
\begin{itemize}
  \item Development of an attention-based convolutional variational autoencoder (C-VAE) for learning
    compact, domain-specific latent representations of large-scale X-ray scattering datasets.
  \item Large-scale latent representation learning from 1.5~million scattering images collected under
    diverse experimental conditions, demonstrating that the learned latent space captures structured
    scattering regimes, experimental trajectories, and continuous morphological transitions.
  \item Deployment of the pre-trained C-VAE for on-the-fly analysis at two synchrotron facilities, with
    a systematic benchmark against DINOv3 (ViT-7B)~\cite{simeoni2025dinov3} as a general-purpose
    baseline, providing preliminary evidence for the advantage of domain-specific training for scattering data.
  \item Development and comparison of two latent space synthetic scattering image generation strategies:
    UMAP-guided PCA sampling and conditional flow matching. Flow matching achieves superior conditioning
    fidelity while both strategies produce physically realistic on-manifold outputs across diverse
    scattering regimes.
  \item Integration of representation learning and interactive visualization into the \mlex\ platform
    through \lsexplorer, supporting real-time embedding visualization during on-the-fly analysis as
    well as offline analysis of scattering data collected across multiple experimental facilities.
\end{itemize}

\section{Methods}
\label{sec:methods}

\subsection{C-VAE Architecture}
\label{subsec:model_arch}

X-ray scattering images contain rich structural information that varies across samples, measurement
geometries, and experimental conditions. Extracting compact and meaningful representations from such
data requires models that capture both local spatial features and longer-range structural patterns. A
Variational Autoencoder~\cite{kingma2013auto} is employed to learn compact latent representations of
X-ray scattering images. VAEs provide a probabilistic framework that enforces a structured and
continuous latent space, making them well-suited for downstream tasks such as clustering,
visualization, and exploration. The resulting latent space captures dominant structural variations
across experiments and supports analysis of structural relationships between scattering patterns.

The C-VAE is a convolutional architecture augmented with localized self-attention mechanisms.
CNNs~\cite{krizhevsky2012imagenet} are effective for modeling local spatial structure in image data,
while attention mechanisms~\cite{vaswani2017attention} enable modeling of longer-range dependencies and
hierarchical relationships. In the encoder, convolutional down-sampling extracts multiscale spatial
features, while attention modules refine contextual relationships within local neighborhoods. These
window-based attention blocks use a window size of w = 8, inspired by Swin-style windowed
attention~\cite{liu2021swin,dosovitskiy2020image}, and allow the model to capture spatial correlations
efficiently while remaining computationally efficient for large images. This hybrid design is particularly well-suited for scattering data, where both localized features and global structural patterns are important. While convolutional layers handle local features efficiently, they struggle to capture broader structural relationships. The windowed self-attention blocks fill that gap by letting each spatial token attend to its neighbors directly.

\begin{figure}[!htbp]
  \centering
  \includegraphics[width=\textwidth,frame]{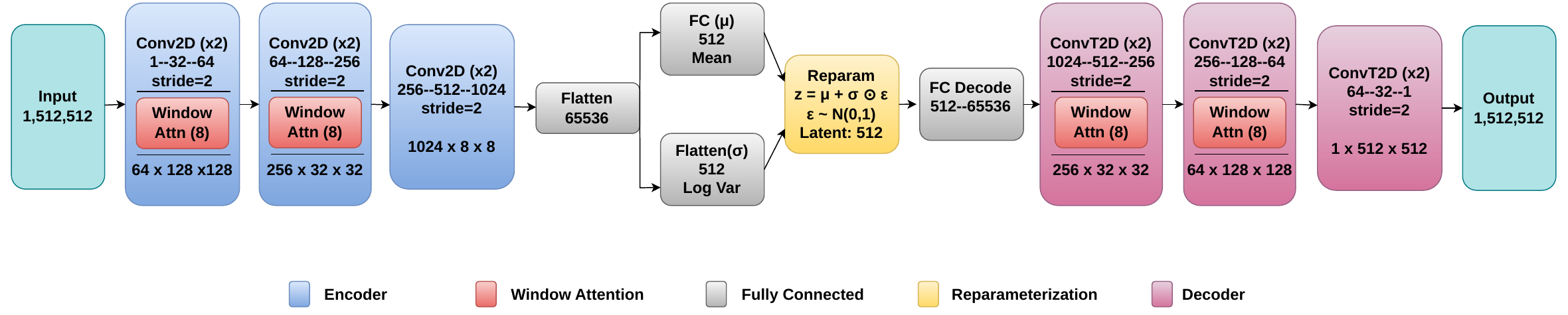}
  \caption{Architecture of the convolutional VAE with windowed attention blocks. Conv2D/ConvT2D denote
    strided and transposed convolution; FC denotes fully connected layers.}
  \label{fig:vae_arch}
\end{figure}

Given an input image $x \in \mathbb{R}^{1 \times H \times W}$, the encoder consists of a hierarchy of
strided convolution layers interleaved with localized attention blocks. Each attention block partitions
the feature map into non-overlapping windows of size $w \times w$ and applies self-attention
independently within each window. For a window $X_w \in \mathbb{R}^{N \times d}$, where $N = w^2$ is
the number of tokens in the window and $d$ is the feature dimension, self-attention is computed as:
\begin{equation}
  \mathrm{Attn}(X_w) = \mathrm{softmax}\!\left(\frac{Q_w K_w^\top}{\sqrt{d_k}}\right) V_w,
\end{equation}
where
\begin{equation}
  Q_w = X_w W_q, \qquad K_w = X_w W_k, \qquad V_w = X_w W_v
\end{equation}
and $W_q, W_k \in \mathbb{R}^{d \times d_k}$, $W_v \in \mathbb{R}^{d \times d_v}$ are learnable
projection matrices. The encoded feature tensor is flattened and mapped to the mean and log-variance of
a diagonal Gaussian latent distribution. Sampling in the latent space is performed using the
reparameterization trick:
\begin{equation}
  z = \mu + \sigma \odot \varepsilon, \qquad
  \sigma = \exp\!\left(\frac{1}{2}\log \sigma^2\right), \qquad
  \varepsilon \sim \mathcal{N}(0, I),
\end{equation}
where $z$ is the sampled latent vector, $\mu$ and $\sigma$ are the mean and standard deviation of the
latent distribution, $\varepsilon$ is a random variable drawn from a standard normal distribution, and
$\odot$ denotes element-wise multiplication. This operation enables backpropagation through stochastic
sampling by expressing the latent variable as a deterministic function of $\mu$, $\sigma$, and random
noise. The decoder mirrors the encoder through transposed convolution layers and symmetric
windowed-attention blocks, progressively reconstructing spatial structure from the latent vector. The
overall model architecture is shown in Figure~\ref{fig:vae_arch}.

The objective is to analyze the learned latent representations rather than to maximize reconstruction
fidelity. Although the decoder reconstructs scattering images from the latent vector, the encoder
embeddings provide the main representation used for downstream analysis. These latent embeddings can be
used for clustering, visualization, and exploration of relationships between scattering patterns.

\subsection{Model Training}
\label{subsec:model_training}

This study uses a dataset of approximately 1.5~million X-ray scattering images collected over the past
decade at the Advanced Light Source (ALS). Each image has a resolution of $1475 \times 1679$ pixels and
is drawn from a randomized subset of a larger archive containing more than two million images acquired
at the SAXS/WAXS beamline 7.3.3~\cite{hexemer2010saxs}. Prior to model input, images are resized to
$512 \times 512$ pixels to match the fixed input dimensions of the C-VAE architecture. The dataset
spans a wide range of material systems including crystalline powders, liquid crystals, amorphous
materials, and thin films, measured in transmission and grazing-incidence geometries. All images were
recorded using a PILATUS3 2M detector composed of 24 modules arranged in three columns of eight
vertically stacked units. Due to the diversity of samples and experimental conditions, the dataset
contains a broad spectrum of scattering signatures, including isotropic rings, sharp diffraction peaks,
streaked patterns, and other complex structures associated with different forms of material
organization.

Model training was performed on the National Energy Research Scientific Computing Center (NERSC)
Perlmutter supercomputer~\cite{nersc_perlmutter_architecture}. We used 20 compute nodes, each equipped
with four NVIDIA A100 GPUs (80~GB), for a total of 80~GPUs. Distributed training was implemented using
\pytorch's Distributed Data Parallel
(DDP)\footnote{\url{https://docs.pytorch.org/tutorials/intermediate/ddp_tutorial.html}} framework to
synchronize model updates across GPUs and enable efficient multi-node scaling. Job scheduling and
resource allocation were managed using the Slurm workload manager. Under this configuration, training
on 1.5~million images required approximately 48~hours.

All models were trained using the Adam optimizer together with a cosine annealing learning rate
scheduler. Input images were normalized prior to training, and distributed data samplers were used to
ensure balanced workloads across GPUs. To improve computational efficiency and reduce GPU memory usage,
automatic mixed precision (AMP) was enabled during training. Additional training details, including
hyperparameter settings and architecture specifications, are provided in Supplementary Note~3.

\subsection{Latent Manifold Sampling for Synthetic Data Generation}
\label{subsec:latent_sampling}

The trained C-VAE decoder can generate scattering images from arbitrary points in the learned latent
space, enabling controlled sampling of the scattering phase space. Two complementary strategies are
implemented for this purpose, both conditioned on a two-dimensional UMAP coordinate and producing a
512-dimensional latent vector that is subsequently decoded by the trained C-VAE decoder to yield a
synthetic scattering image.

In the first strategy, UMAP-guided PCA sampling, cluster-aware PCA models are fitted to the subset of
core latent vectors whose HDBSCAN cluster membership strength exceeds 0.8. For each cluster, the
centroid is computed in the two-dimensional UMAP embedding using these core samples. To generate a new
sample, Euclidean distances from a query UMAP coordinate to all cluster centroids are computed and
converted into a probability distribution over clusters using a temperature-controlled top-$k$ softmax
weighting scheme, which assigns higher probability to nearby clusters while allowing limited
contributions from neighbouring clusters. In our implementation we use $k = 2$ and a temperature of
$T = 0.1$. Intra-cluster variability is modelled using PCA fitted to the core latent vectors of each
cluster. New latent candidates are sampled by drawing Gaussian noise in the PCA coordinate space and
inverse-transforming to the full latent space. To reduce out-of-distribution artefacts, per-dimension
clamping is applied, restricting each latent dimension to the interval $\mu \pm 2.5\sigma$, where $\mu$
and $\sigma$ denote the global mean and standard deviation of the core latent vectors. Full pipeline
details are provided in Supplementary Note~2.1.

In the second strategy, conditional flow matching, a dedicated velocity network is trained to learn a
continuous normalizing flow~\cite{lipman2022flow} from a standard Gaussian prior
$\mathcal{N}(\mathbf{0}, \mathbf{I})$ to the training latent distribution, conditioned on the
two-dimensional UMAP coordinate of the desired scattering state. The velocity network consists of six
residual blocks with adaptive layer normalization, conditioned on a fused embedding of a sinusoidal
time encoding and a learned UMAP encoder, both projected to a 256-dimensional conditioning space. To
support conditioning strength control at inference, classifier-free
guidance~\cite{ho2022classifier} is incorporated during training by randomly zeroing the UMAP
conditioning with probability $p = 0.15$, training the network simultaneously as a conditional and
unconditional model. At inference, a guided velocity is constructed as a weighted combination of
conditional and unconditional predictions, with guidance scale $s = 5.0$, and the resulting ordinary
differential equation is integrated over 20 Euler steps to produce the synthetic latent vector. The
model was trained on high-confidence cluster members with HDBSCAN membership strength exceeding 0.5,
using the AdamW optimiser with a learning rate of $3 \times 10^{-4}$ and weight decay of $10^{-4}$,
combined with a cosine annealing learning rate scheduler. Full architecture and training details are
provided in Supplementary Note~2.2.

\subsection{Software Infrastructure}
\label{subsec:software}

\begin{figure}[h]
  \centering
  \includegraphics[width=0.92\textwidth,frame]{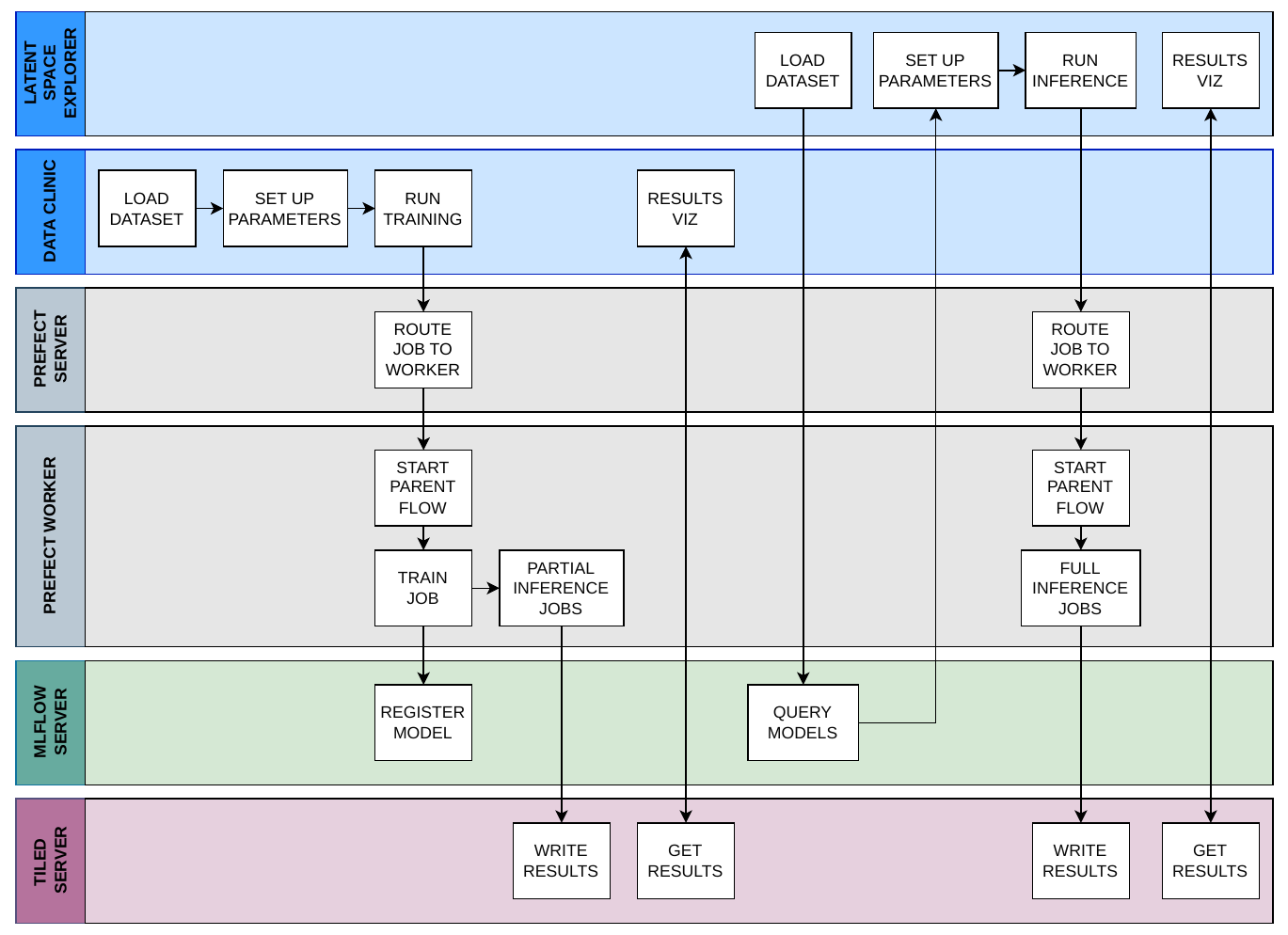}
  \caption{Software architecture diagram of \protect\lsexplorer\ within the \mlex\ platform. The
    system integrates workflow orchestration (\protect\prefect), data management (\protect\tiled), and
    model tracking (\protect\mlflow, \protect\dataclinic).}
  \label{fig:lse_diagram}
\end{figure}

The \lsexplorer\ application is integrated within the \mlex\ platform and shares infrastructure with
several web-based applications for workflow orchestration, data management, and visualization as shown
in Figure~\ref{fig:lse_diagram}. It extends interactive latent space exploration to large-scale X-ray
scattering datasets, integrating domain-specific model training, distributed workflow execution, and
real-time experimental data streams within a unified environment.

\subsubsection*{Data access}

\lsexplorer\ supports data access from both the local file system and
\tiled\footnote{\url{https://blueskyproject.io/tiled/}}. \tiled\ is a data access service within the
\bluesky\ ecosystem~\cite{allan2019bluesky} that provides HTTP-based access to scientific datasets. It
is widely used at synchrotron facilities to manage and serve experimental data generated during beamline
operations. Data loading from these sources is enabled through an internally developed tool called
\filemanager\footnote{\url{https://github.com/mlexchange/mlex_file_manager}}~\cite{chavez2025machine},
which provides a unified interface for seamless data integration across these sources. File system
access is currently restricted to common image formats including PNG, JPEG, and TIF files. In contrast,
\tiled\ supports a broader range of dataset types through native compatibility and customizable
ingestion mechanisms, allowing it to accommodate structured experimental data produced by beamline
instruments.

For results data, \lsexplorer\ writes and retrieves outputs exclusively through \tiled. Generated
results are cataloged using a human-readable hashed identifier derived from the dataset name, enabling
the system to handle multiple datasets or directories consistently.

\subsubsection*{Job management}

Formerly, the execution of ML jobs was managed by \mlexcompute, an internally developed orchestration
system within \mlex~\cite{zhao2022mlexchange,chavez2025machine}. This system has since been superseded
by a \prefect\footnote{\url{https://docs.prefect.io/v3/get-started}} based workflow orchestration.
\prefect\ is a workflow management system that enables distributed execution of computational tasks
while maintaining clear dependencies between processing stages. In the current setup, ML job requests
are handled through a central (parent) \prefect\ flow that packages the job according to
user-specified parameters. Depending on the task, this flow may trigger multiple subordinate workflows
(subflows). For example, during model training two subflows are typically executed: one responsible for
the training stage and another responsible for performing partial inference on a subset of the dataset.
Job execution is carried out by \prefect\ workers, which support multiple execution backends including
Docker, Podman, Slurm, and Conda environments.

\subsubsection*{User interface}

\begin{figure}[!htbp]
  \centering
  \begin{subfigure}[t]{\linewidth}
    \centering
    \includegraphics[width=\textwidth, frame]{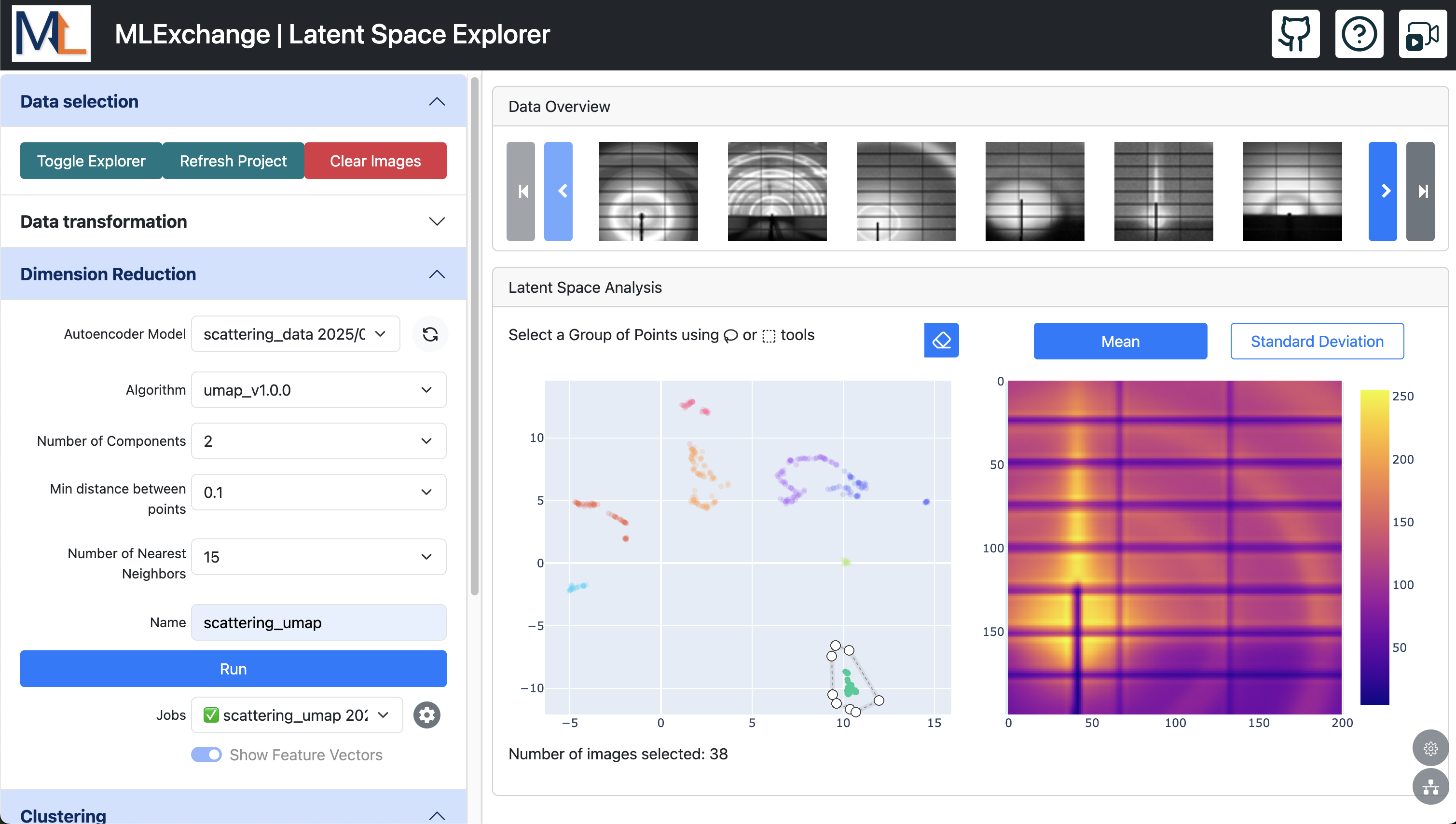}
    \caption{User interface of \protect\lsexplorer. The left sidebar enables users to select datasets,
      configure parameters, and initiate dimensionality reduction and clustering algorithms. The right
      panel presents an overview of the selected dataset and supports interactive exploration of the
      latent space, including the visualization of statistical summaries.}
    \label{fig:lse_frontend}
  \end{subfigure}
  \vspace{1em}
  \begin{subfigure}[t]{\linewidth}
    \centering
    \includegraphics[width=\textwidth, frame]{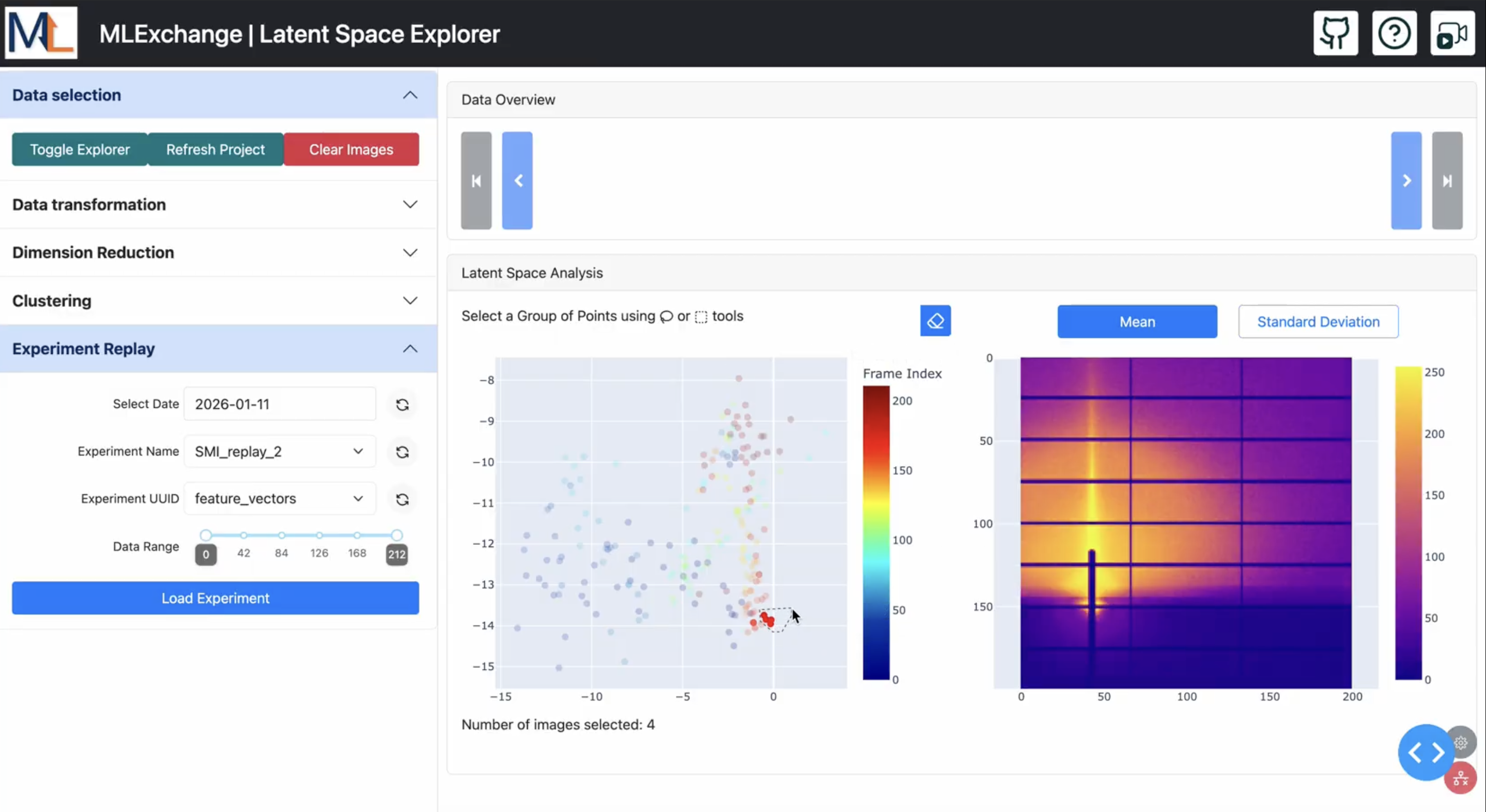}
    \caption{Live mode operation in \protect\lsexplorer. Dataset information and corresponding feature
      vectors are streamed in real time through a websocket connection. Users can visualize and interact
      with regions of interest directly within the plot. Frame index corresponds to the time chronology
      of incoming data. Live mode is activated by clicking the Live button located in the top-right bar
      of the interface.}
    \label{fig:lse_live_mode}
  \end{subfigure}
  \caption{The \protect\lsexplorer\ interface in standard and live modes.}
  \label{fig:lse_interface}
\end{figure}

The \lsexplorer\ application provides an interactive environment for exploring the latent space
representations of scientific datasets, as shown in Figure~\ref{fig:lse_interface}. Through the web
interface, users can load datasets, configure analysis parameters, and execute dimensionality reduction
workflows. The interface supports PCA and UMAP, which can be applied to selected datasets with
configurable parameters. Once dimensionality reduction has been completed, the resulting embeddings can
be visualized in two or three dimensional latent space. Users may interactively select regions of
interest within this space to investigate groups of similar scattering patterns. Statistical summaries
such as the mean and standard deviation of selected data points can then be computed and visualized. In
addition, clustering algorithms such as DBSCAN and HDBSCAN can be applied to identify groups of
structurally related scattering patterns.

Optionally, users may incorporate latent embeddings generated by pretrained autoencoders developed
within the \dataclinic~\cite{chavez2025machine} application of \mlex. \dataclinic\ offers a
web-based interface for interactively training and evaluating tunable autoencoders on scientific
datasets. However, through integration with \mlflow, users are not limited to models developed within
\dataclinic\ and can register and utilize more complex or externally trained models.

\subsubsection*{Model registration}
\label{sec:model_reg}

To enable seamless exchange of trained autoencoders across different components of the \mlex\ platform,
\mlflow\ server\footnote{\url{https://mlflow.org}} is used to register trained models together with
their associated training parameters~\cite{zaharia2018accelerating}. This approach ensures versioned
storage of trained models, enabling reproducibility and consistent reuse across multiple applications.
Once a model has been trained and registered with the ALS \mlflow\
server\footnote{\url{https://mlflow.computing.als.lbl.gov}}, it becomes available within the
\lsexplorer\ interface. Users can select a registered model through a drop-down menu and apply the
corresponding encoder to generate latent embeddings prior to dimensionality reduction or clustering.

\subsubsection*{On-the-fly capabilities}

\lsexplorer\ supports on-the-fly capabilities for real-time data embedding, enabling visualization of
experimental data as it is generated. The inference time for the autoencoder and dimensionality
reduction model is approximately 0.04~sec and 0.005~sec, respectively, resulting in a total processing
time of about 0.05~sec per sample. Additional compute resources can be allocated to scale inference to
faster acquisition rates. To support streaming data processing, we use
\arroyo,\footnote{\url{https://pypi.org/project/arroyopy/}} a framework that integrates with messaging
systems and supports flexible composition of processing steps. ML models are executed directly within
the streaming pipeline, and the resulting latent embeddings are projected using a pre-fitted
dimensionality reduction model before being transmitted to the frontend.

From the perspective of the web interface, a websocket connection continuously listens for incoming
messages containing a \tiled\ URI together with the corresponding two-dimensional feature vector. These
vectors are dynamically aggregated and rendered in the frontend, enabling users to observe the evolving
latent space in real time. Users can interact directly with this visualization by selecting regions of
interest and computing statistical summaries. In this real-time workflow, data retrieval is performed
exclusively through the \tiled\ data service, ensuring consistent access to experimental datasets
generated during beamline operations.

\section{Results}
\label{sec:results}

The results are organized to reflect the two deployment settings introduced in
Section~\ref{sec:intro}. We first characterize the latent space learned from the historical training
dataset (Section~\ref{subsec:explore}), demonstrating that the C-VAE produces a well-organized
representation across 1.5~million scattering images. We then deploy this pre-trained model for
on-the-fly analysis of previously unseen scattering data collected during live experiments at two
synchrotron facilities (Section~\ref{subsec:otf}), and benchmark its performance against DINOv3
(ViT-7B)~\cite{simeoni2025dinov3} as a general-purpose baseline to evaluate the benefit of
domain-specific training. Finally, we demonstrate the generative capabilities of the learned
representation through UMAP-guided latent sampling (Section~\ref{subsec:data_generation}). All
detector images are displayed using a plasma colormap for visualization, although the underlying data
are grayscale intensity images.

Additionally, \lsexplorer\ was deployed for post-experiment data analysis at the P03 micro- and
nanofocus small- and wide-angle X-ray scattering (MiNaXS) beamline at ALS 9.3.1 beamline and
PETRA~III (DESY, Germany); while highly promising, these results are beyond the scope of this
methods-focused study and will be reported separately in a science-driven publication.

\subsection{Exploring X-ray Scattering Data}
\label{subsec:explore}

\begin{figure}[!htbp]
\centering
\begin{mdframed}[linewidth=0.8pt, linecolor=black, innerleftmargin=8pt, innerrightmargin=8pt,
                 innertopmargin=8pt, innerbottommargin=8pt]
\begin{subfigure}[t]{0.30\linewidth}
  \centering
  \includegraphics[width=\linewidth]{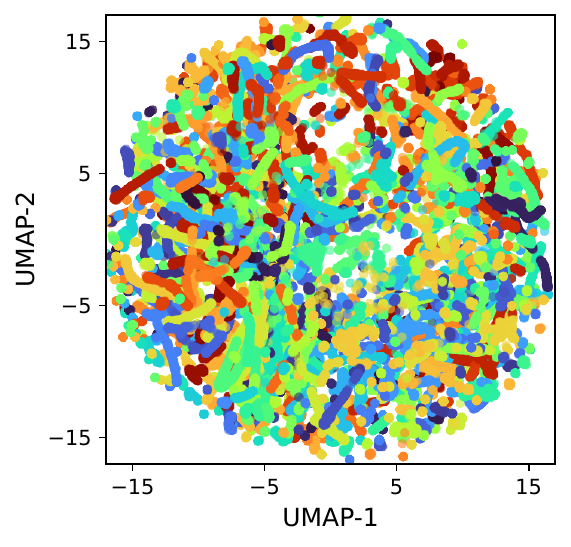}
  \caption{1.5M C-VAE embeddings projected via UMAP, colored by HDBSCAN cluster assignment.}
  \label{fig:density}
\end{subfigure}
\hfill
\begin{subfigure}[t]{0.30\linewidth}
  \centering
  \includegraphics[width=\linewidth]{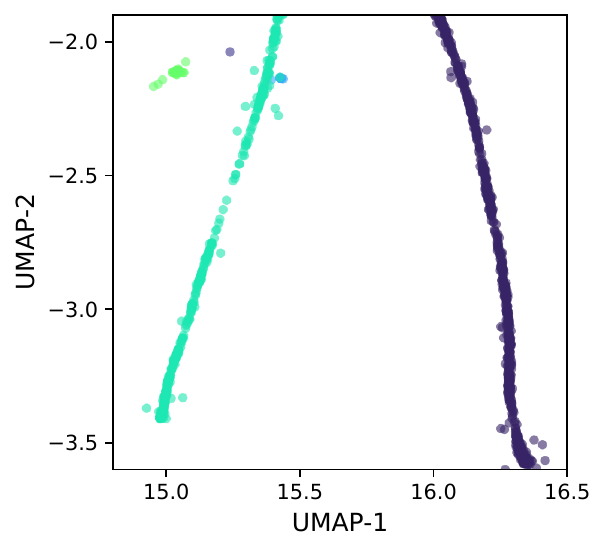}
  \caption{Zoomed view of two well-separated experiment clusters in C-VAE latent space.}
  \label{fig:multi_clusters}
\end{subfigure}
\hfill
\begin{subfigure}[t]{0.30\linewidth}
  \centering
  \includegraphics[width=\linewidth]{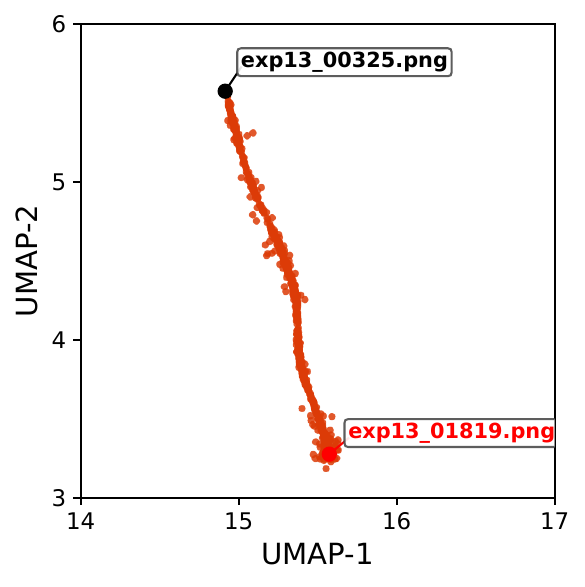}
  \caption{Single-cluster trajectory reflecting temporal progression within one experiment.}
  \label{fig:single_cluster}
\end{subfigure}
\end{mdframed}
\caption{Latent space structure of C-VAE embeddings projected via UMAP. \textbf{(a)}~The full training
  dataset exhibits a continuous yet structured manifold, with HDBSCAN clusters corresponding to groups
  of structurally similar scattering patterns. \textbf{(b)}~Independent experiments form well-separated
  clusters in the latent space. \textbf{(c)}~Within a single cluster, embeddings follow a smooth
  trajectory reflecting the temporal progression of the experiment.}
\label{fig:latent_analysis}
\end{figure}

We begin by analyzing the structure of the training dataset used to train the C-VAE model. The dataset
contains approximately 1.5~million X-ray scattering images collected under diverse experimental
conditions (see Methods). The images were stored in RGBA format with dimensions of $1475 \times 1679$
pixels and were resized to $512 \times 512$ prior to encoding with the C-VAE model. The encoder
generated a 512-dimensional latent representation for each image, providing a compact description of
the structural features present in the scattering patterns. To visualize the global structure of this
high-dimensional latent space, the embeddings were projected into two dimensions using
UMAP~\cite{mcinnes2018umap}. The resulting visualization revealed a highly non-uniform distribution of
points, reflecting the diversity of scattering patterns present in the dataset. To identify recurring
structural motifs, we applied the HDBSCAN clustering algorithm~\cite{campello2013density} to the latent
space. Unlike $k$-means clustering, HDBSCAN automatically determines the number of clusters and
identifies ambiguous or noisy samples as outliers. Figure~\ref{fig:latent_analysis} shows the UMAP
projection of the full training dataset together with representative clusters. The visualization
revealed a continuous yet structured latent space in which clusters corresponded to groups of
scattering patterns with similar structural characteristics, an organization that could not be recovered
by projecting raw pixel intensities directly through UMAP without using a trained encoder
(Supplementary Note~1.2). As illustrated in Figure~\ref{fig:multi_clusters}, independent experiments
formed well-separated clusters, indicating that the latent representation encoded experiment-specific
information. Notably, neighboring clusters remained close in latent space, reflecting similarities
between related experiments. In addition, Figure~\ref{fig:single_cluster} shows that each cluster
followed a line-like trajectory corresponding to the temporal evolution of individual experiments,
demonstrating that the latent space preserved both structural similarity and experimental progression. A
PCA analysis of the 512-dimensional latent vectors is provided in Supplementary Note~1.1.

\begin{figure}[!htbp]
\centering
\begin{mdframed}[linewidth=0.6pt, linecolor=black, innerleftmargin=8pt, innerrightmargin=8pt,
                 innertopmargin=8pt, innerbottommargin=8pt]
\begin{subfigure}[t]{0.46\linewidth}
  \centering
  \includegraphics[width=\linewidth, height=4cm]{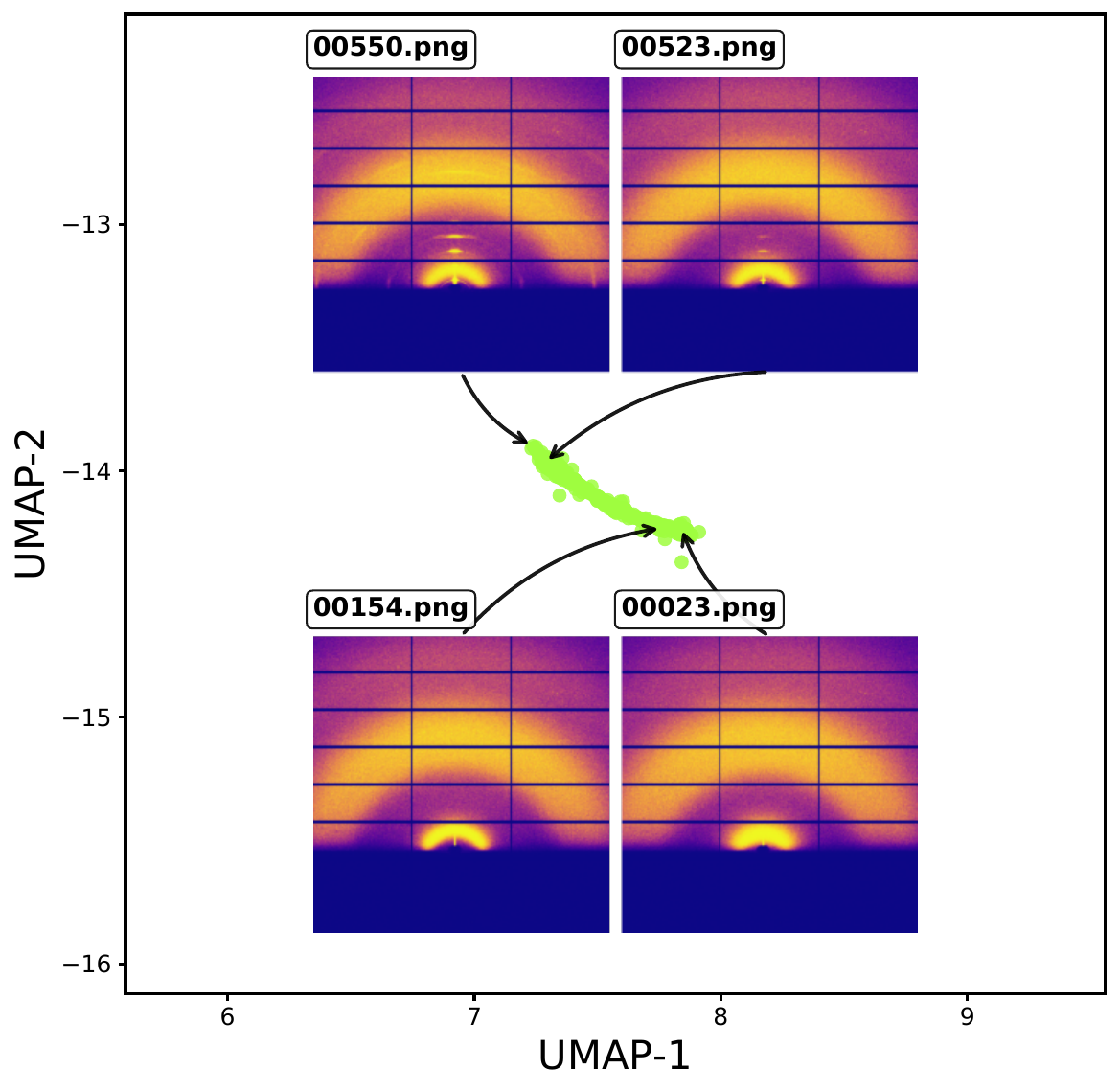}
  \caption{UMAP trajectory of an experiment with representative scattering images overlaid.}
  \label{fig:exp_progress}
\end{subfigure}
\hfill
\begin{subfigure}[t]{0.46\linewidth}
  \centering
  \includegraphics[width=\linewidth,height=4cm]{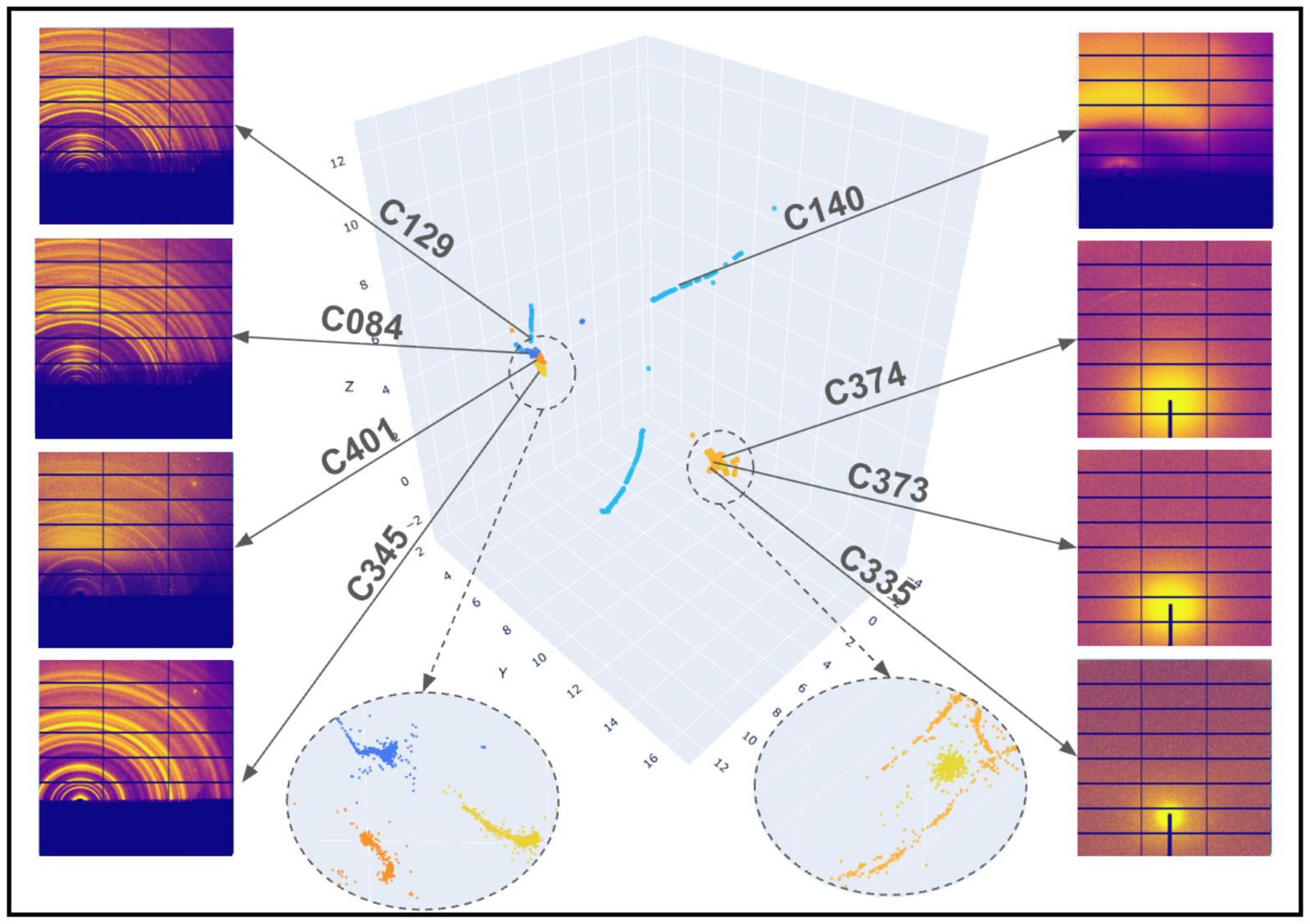}
  \caption{3D cluster view revealing similarity and dissimilarity of clusters in latent space.}
  \label{fig:3d_zoom}
\end{subfigure}
\end{mdframed}
\caption{Complementary views of C-VAE latent structure. \textbf{(a)}~UMAP projection of a single
  experimental run, with representative scattering images overlaid at selected frames to illustrate how
  structural features evolve along the latent trajectory. \textbf{(b)}~Three-dimensional visualization
  of latent clusters, showing separation between groups that appear overlapping in two-dimensional
  projections.}
\label{fig:latent_visualizations}
\end{figure}

Figure~\ref{fig:latent_visualizations} shows two complementary views of the latent structure.
Figure~\ref{fig:exp_progress} illustrates the progression of a scattering experiment within the learned
latent space. The 512-dimensional embeddings are projected into two dimensions using UMAP, where each
point corresponds to a single scattering image. Representative images along the trajectory reveal a
smooth transition between scattering patterns. For example, the initial image (00023) exhibits a
semicircular scattering pattern, while subsequent images gradually introduce additional structural
features such as localized intensity variations and concentric rings. The smooth trajectory in latent
space indicates that the C-VAE captures gradual and meaningful variations in the scattering patterns.

While two-dimensional visualizations provide an intuitive overview of the latent structure,
dimensionality reduction may introduce apparent overlap between clusters that are well separated in
higher dimensions. To illustrate this effect, Figure~\ref{fig:3d_zoom} presents a three-dimensional
visualization of the latent clusters. In this representation, clusters that appear overlapping in
two-dimensional projections become clearly separated along additional axes. For example, clusters C084,
C345, and C401 that appear closely spaced in the two-dimensional embedding are more clearly
distinguished in three dimensions. Similarly, clusters C335, C373, and C374 become separable when
additional latent dimensions are considered. This observation highlights the importance of analyzing the
latent structure beyond low-dimensional projections.

The structured latent space learned by the C-VAE provides a compact and interpretable organization of
large scattering datasets, in which clusters correspond to distinct scattering regimes and trajectories
reflect the temporal progression of individual experiments. These results confirm that training on
1.5~million historical scattering images produces a latent space that captures physically meaningful
structure across diverse experimental conditions. This pre-trained model is therefore well-suited for
deployment in on-the-fly analysis of new experiments, without any retraining.

\subsection{On-the-Fly Analysis}
\label{subsec:otf}

Having established that the C-VAE learns a well-organized latent space from historical training data,
we now evaluate its deployment for on-the-fly analysis of previously unseen experiments. We consider
time-resolved grazing-incidence X-ray scattering measurements of perfluorosulfonic acid (PFSA, Nafion)
ionomer films, performed to probe the morphological evolution of PFSA ionomer films during blade coating
and drying~\cite{dudenas2019evolution,kusoglu2017new}. Sequential scattering images were acquired
throughout film formation, capturing transitions from solvent-dominated states to semicrystalline
aggregates and fully formed structures. The trained C-VAE model is applied without retraining to data
acquired in on-the-fly mode at beamline 7.3.3 at the Advanced Light Source (Case Study~1) and at the
Soft Matter Interfaces (SMI) beamline at NSLS-II (Case Study~2). In this setting, detector images
captured during the experiment are streamed directly into the analysis pipeline through a websocket
listener. Each image is embedded into the learned latent space in real time and visualized through the
\lsexplorer\ interface, enabling immediate inspection of latent trajectories and clustering behavior
without requiring offline post-processing. Each captured high-resolution image ($1475 \times 1679$) is
first resized to $512\times512$ and then encoded into a $512$-dimensional latent mean vector, which
serves as the representation used for all downstream analysis.

\begin{figure}[p]
\centering
\captionsetup[subfigure]{labelfont=bf, justification=centering, skip=4pt}
\subcaptionbox{ALS beamline 7.3.3 (C-VAE, PCA)\label{fig:inset_als}}{%
  \includegraphics[width=0.82\linewidth, height=0.40\textheight, keepaspectratio]{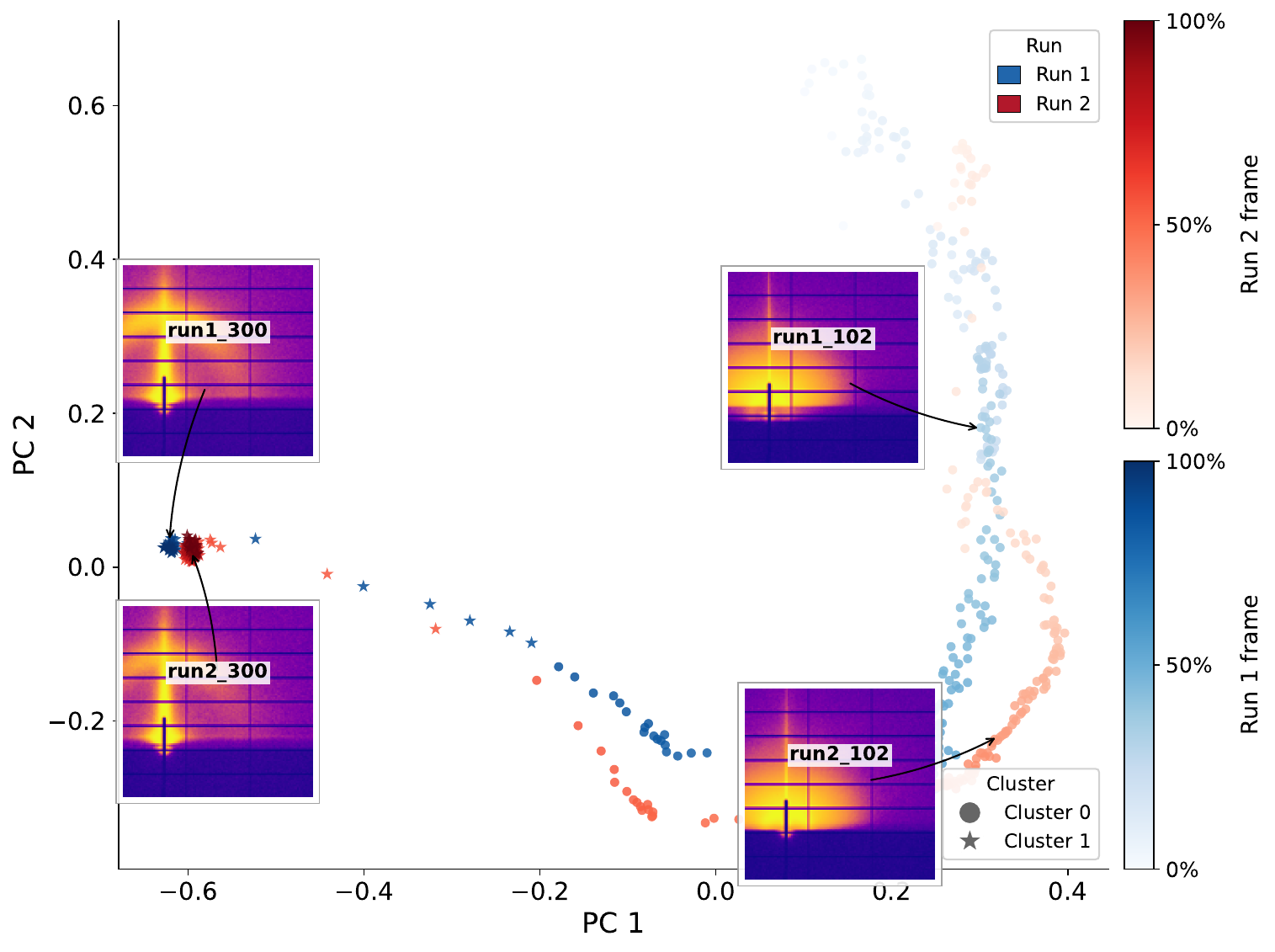}}
\vspace{0.5em}
\subcaptionbox{NSLS-II SMI beamline (C-VAE, PCA)\label{fig:inset_nsls}}{%
  \includegraphics[width=0.82\linewidth, height=0.40\textheight, keepaspectratio]{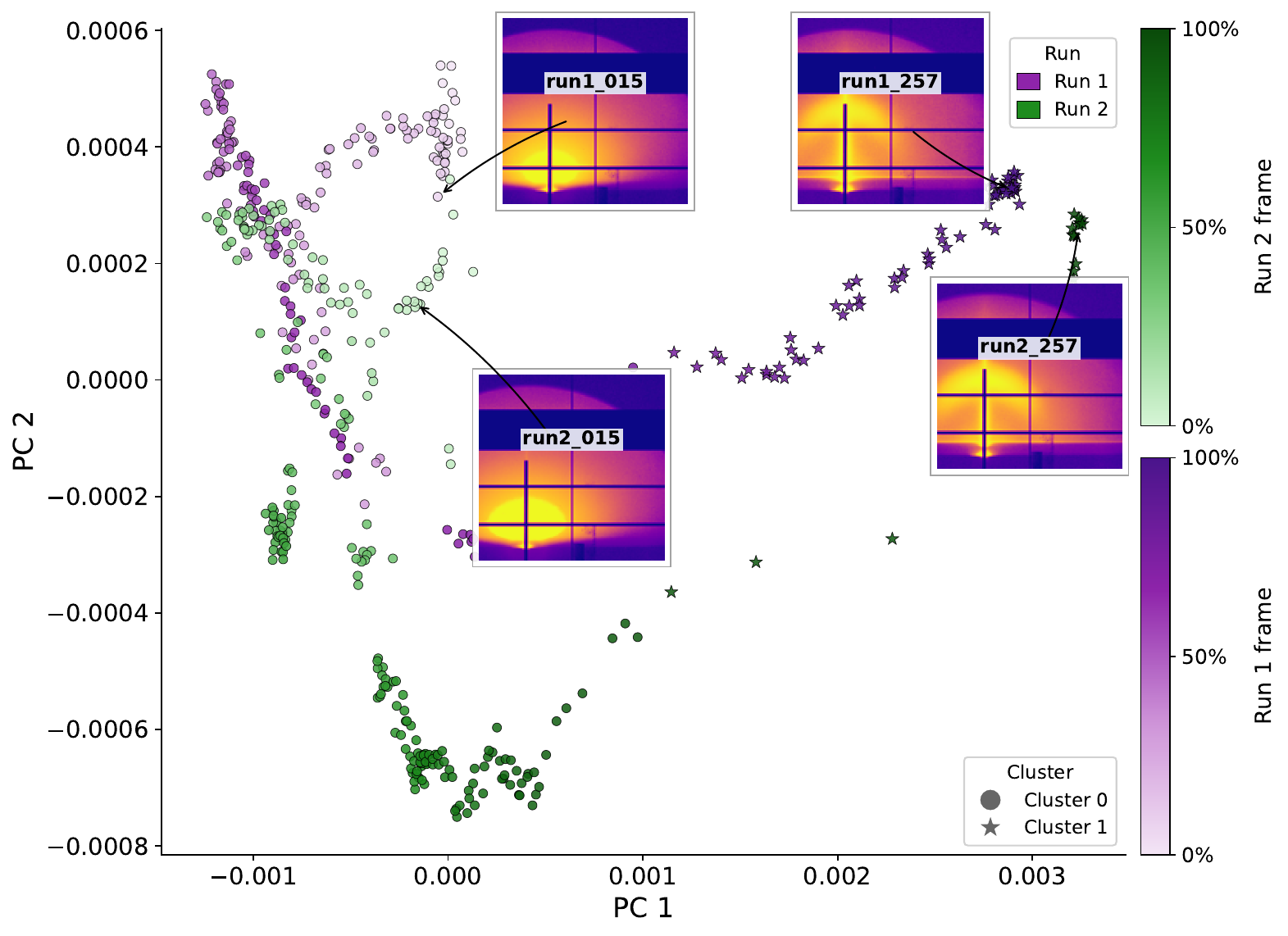}}
\caption{C-VAE PCA trajectories with representative scattering images overlaid at selected frames for
  \textbf{(a)}~ALS beamline 7.3.3 and \textbf{(b)}~NSLS-II SMI beamline. Color encodes temporal
  progression (0\%=start, 100\%=end); marker shape indicates cluster identity (circle: cluster~0,
  star: cluster~1). Image insets link latent positions to the underlying detector patterns, directly
  connecting structural evolution in the scattering data to the trajectory path in latent space.}
\label{fig:cvae_inset_comparison}
\end{figure}

\paragraph{Case Study 1: Real-Time Analysis at the ALS 7.3.3 Beamline}
The trained C-VAE model is applied to two experimental runs of PFSA ionomer film formation at ALS
beamline 7.3.3, producing smooth latent trajectories that track the structural evolution of the sample
throughout the drying process. Figure~\ref{fig:cvae_inset_comparison}a shows the C-VAE PCA trajectory
with representative scattering images overlaid at selected frames. At early time points, when the
system is in its dispersion state, the scattering pattern is dominated by a diffuse halo at low $q$
(scattering vector or momentum transfer), reflecting solvent scattering. As the dispersion dries, the
polymer begins to form semi-crystalline aggregates, producing an arch-shaped feature within the initial
solvent scattering peak. Once most of the solvent has evaporated, the system self-assembles and
hydrophilic domains form throughout the material, generating a new scattering feature at higher $q$.
The smooth trajectory in latent space directly tracks this physical evolution, confirming that the
C-VAE captures meaningful structural transitions during film formation.

\begin{figure}[p]
\centering
\captionsetup[subfigure]{labelfont=bf, justification=centering, skip=4pt}
\subcaptionbox{ALS beamline 7.3.3: C-VAE (top row) vs.\ DINOv3 (bottom row)%
               \label{fig:traj_als}}{%
  \includegraphics[width=\linewidth, height=0.38\textheight, keepaspectratio]{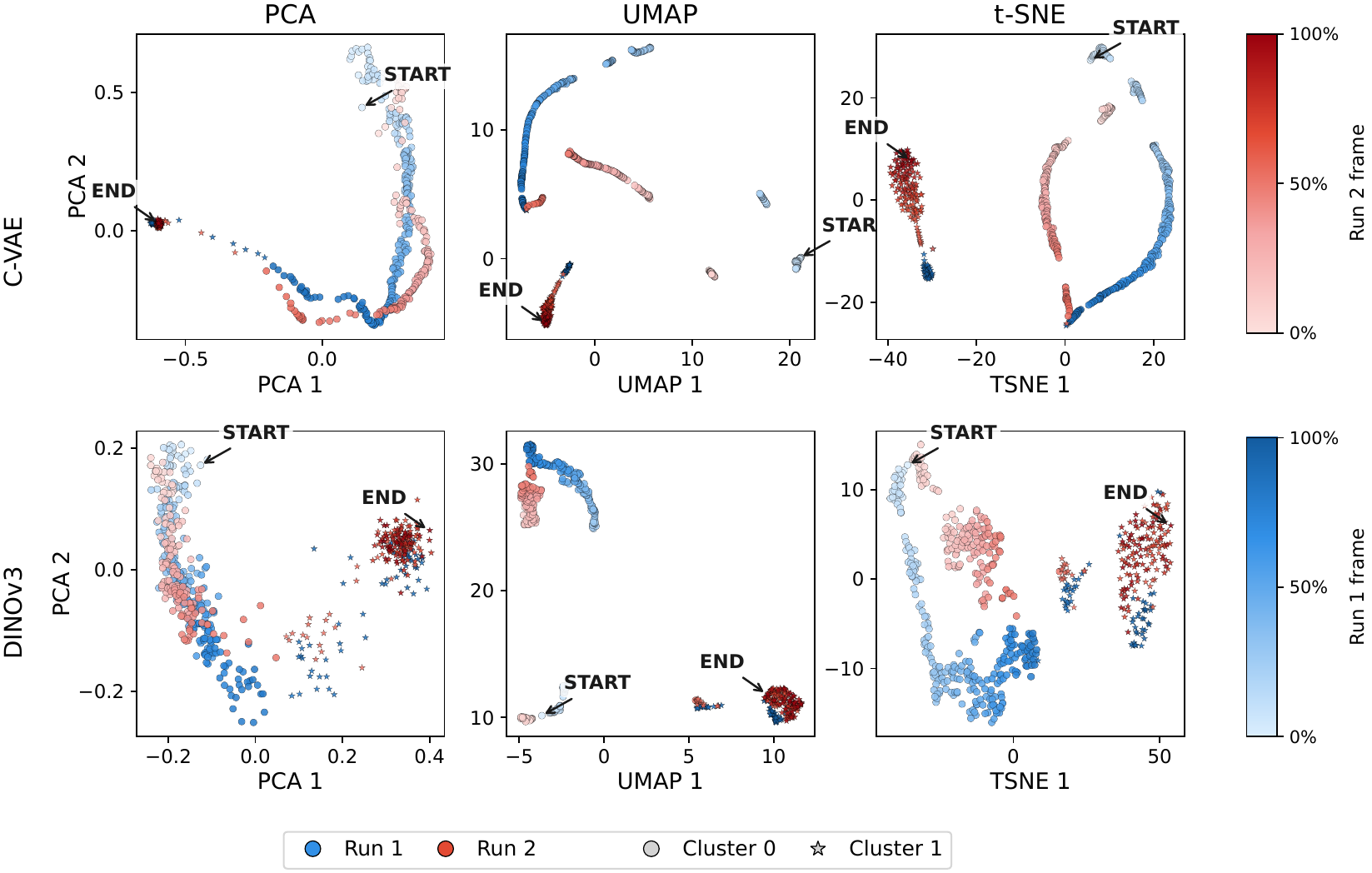}}
\vspace{0.3em}
\subcaptionbox{NSLS-II SMI beamline: C-VAE (top row) vs.\ DINOv3 (bottom row)%
               \label{fig:traj_nsls}}{%
  \includegraphics[width=\linewidth, height=0.38\textheight, keepaspectratio]{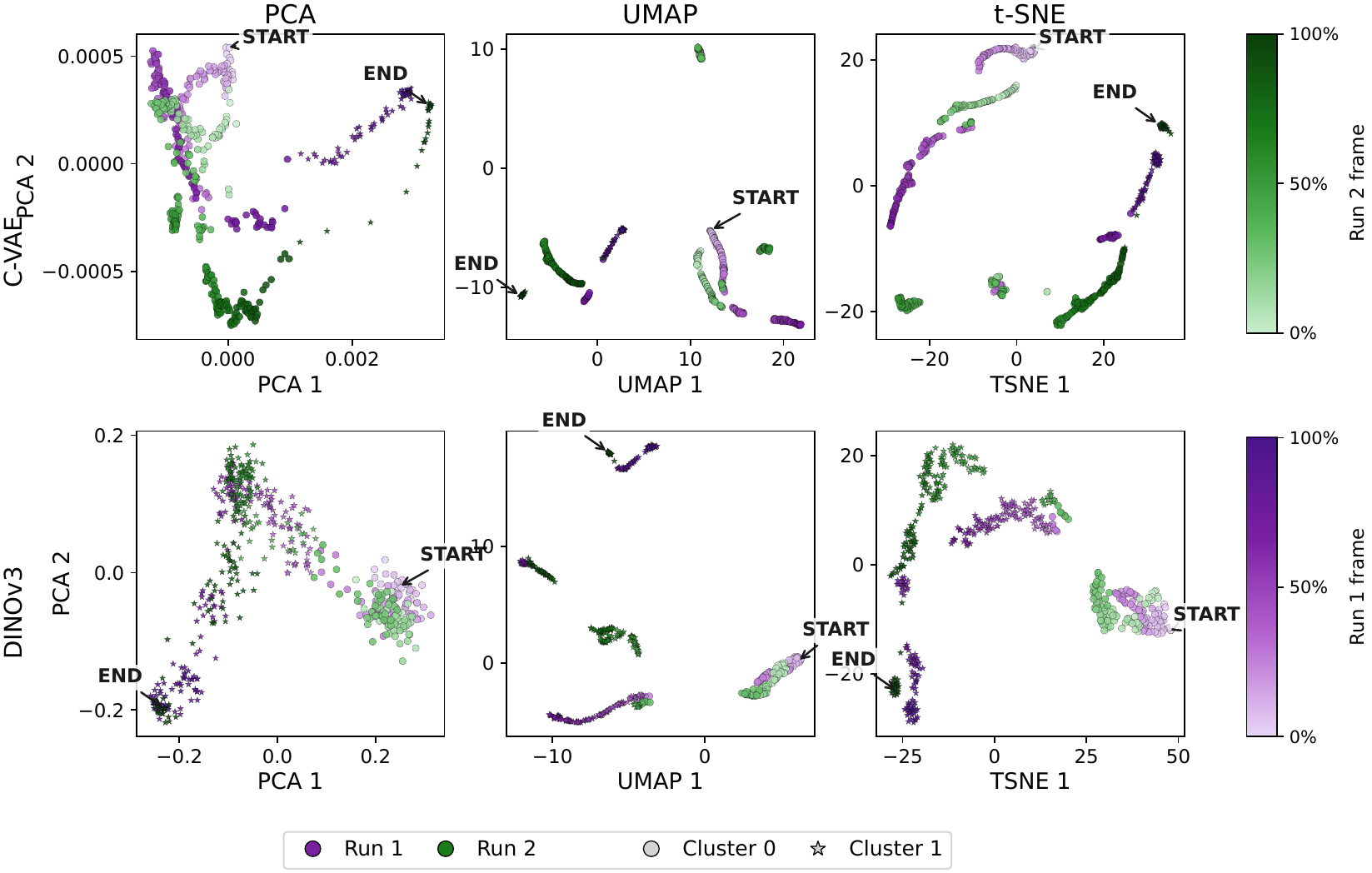}}
\caption{Latent-space trajectories of PFSA ionomer film formation at
  \textbf{(a)}~ALS beamline 7.3.3 and \textbf{(b)}~NSLS-II SMI beamline, comparing C-VAE (top row)
  and DINOv3 (bottom row) embeddings projected via PCA (left), UMAP (centre), and t-SNE (right). Point
  colour encodes temporal frame index within each run (colorbars); marker shape indicates cluster
  identity (circle: cluster~0, star: cluster~1); hue family distinguishes experimental runs.
  \textbf{START} and \textbf{END} mark the global first and last frames of each acquisition
  trajectory.}
\label{fig:combined_trajectory_comparison_als}
\end{figure}

The full comparison across PCA, UMAP, and t-SNE projections is shown in
Figure~\ref{fig:combined_trajectory_comparison_als}a (top row). Each point corresponds to a detector
image acquired sequentially during the experiment, with color encoding the frame index within each run.
Across all projections, the scans form smooth trajectories in latent space, indicating that the model
captures gradual structural evolution occurring during the film formation. The two runs follow similar
but distinct trajectories, suggesting that the representation captures consistent experimental
progression while preserving run-specific characteristics. Cluster assignments obtained through
unsupervised clustering are indicated by marker shape; the kinetically active stage of the sample in
both runs is grouped together (circles) and the kinetically stable stage is grouped together (stars).

\paragraph{Case Study 2: Generalization to NSLS-II Experiments}

Having established that the C-VAE captures consistent latent structure at the ALS, we next ask whether
these representations transfer to a different facility with distinct detector hardware and beamline
geometry. To evaluate this, the trained C-VAE model is applied without retraining to PFSA ionomer film
formation experiments at the NSLS-II SMI beamline, a setting the model had not encountered during
training.  Figure~\ref{fig:cvae_inset_comparison}b shows the C-VAE PCA trajectories with representative
detector images overlaid at selected frames, linking latent positions to the underlying scattering
patterns.

The C-VAE latent trajectories across all projections are shown in
Figure~\ref{fig:combined_trajectory_comparison_als}b (top row). As in the ALS experiments, the scans
form smooth trajectories in latent space, confirming that the model captures the continuous structural
evolution of scattering patterns at a different facility and detector configuration. The separation
between runs indicates that the representation distinguishes between different experimental conditions
or progression stages. Cluster assignments remain well separated across PCA, UMAP, and t-SNE
projections, grouping scans with similar structural features into distinct regions along the
trajectories.

\paragraph{Comparison with a General-Purpose Vision Model}

To evaluate the benefit of domain-specific training and directly test the hypothesis raised in
Section~\ref{sec:intro}, we benchmark the C-VAE against DINOv3
(ViT-7B)~\cite{simeoni2025dinov3}, a large self-supervised vision transformer trained on natural images. As described in Section 1, DINOv3 (ViT-7B) was selected as the general-purpose baseline because its self-supervised ViT architecture most closely mirrors the C-VAE, ensuring that performance differences reflect training data rather than architectural choices, and because its seven billion parameter scale makes any demonstrated advantage conservative. Other widely used foundation models are less suited to this comparison, as language-supervised and segmentation-oriented architectures lack a natural correspondence to the global structural representation required for scattering data.

Embeddings from DINOv3 are computed for the same scattering images used in both case studies
and projected via PCA, UMAP, and t-SNE for direct comparison with the C-VAE. The results for both
facilities are shown in Figure~\ref{fig:combined_trajectory_comparison_als} (bottom rows of each
panel).


To quantitatively support the visual differences observed in Figure~\ref{fig:combined_trajectory_comparison_als}, we computed trajectory smoothness as the mean cosine distance between consecutive frames and cosine silhouette scores ($k{=}2$) in the high-dimensional  latent space. At ALS beamline 7.3.3, C-VAE trajectories exhibit substantially larger frame-to-frame variation along experimental trajectories than DINOv3 (mean cosine distance 0.097 vs.\ 0.032), indicating that domain-specific model is likely more sensitive to structural transitions during film formation. Cluster separation metrics at ALS are comparable between the two models, reflecting that both encoders distinguish the two kinetic stages but differ in the sensitivity of their latent response to transitions between them. At NSLS-II, C-VAE achieves higher silhouette scores across both runs (Run~1: 0.696 vs.\ 0.561; Run~2: 0.824 vs.\ 0.637), indicating that structural organization generalizes well to the unseen facility. Trajectory smoothness at NSLS-II is near zero for both models and is not a meaningful differentiator in this setting. Whether facility-specific fine-tuning of either model further improves latent organization at unseen beamlines remains an open question and a direction for future work.

\begin{figure}[!htbp]
\centering
\includegraphics[width=\linewidth, frame]{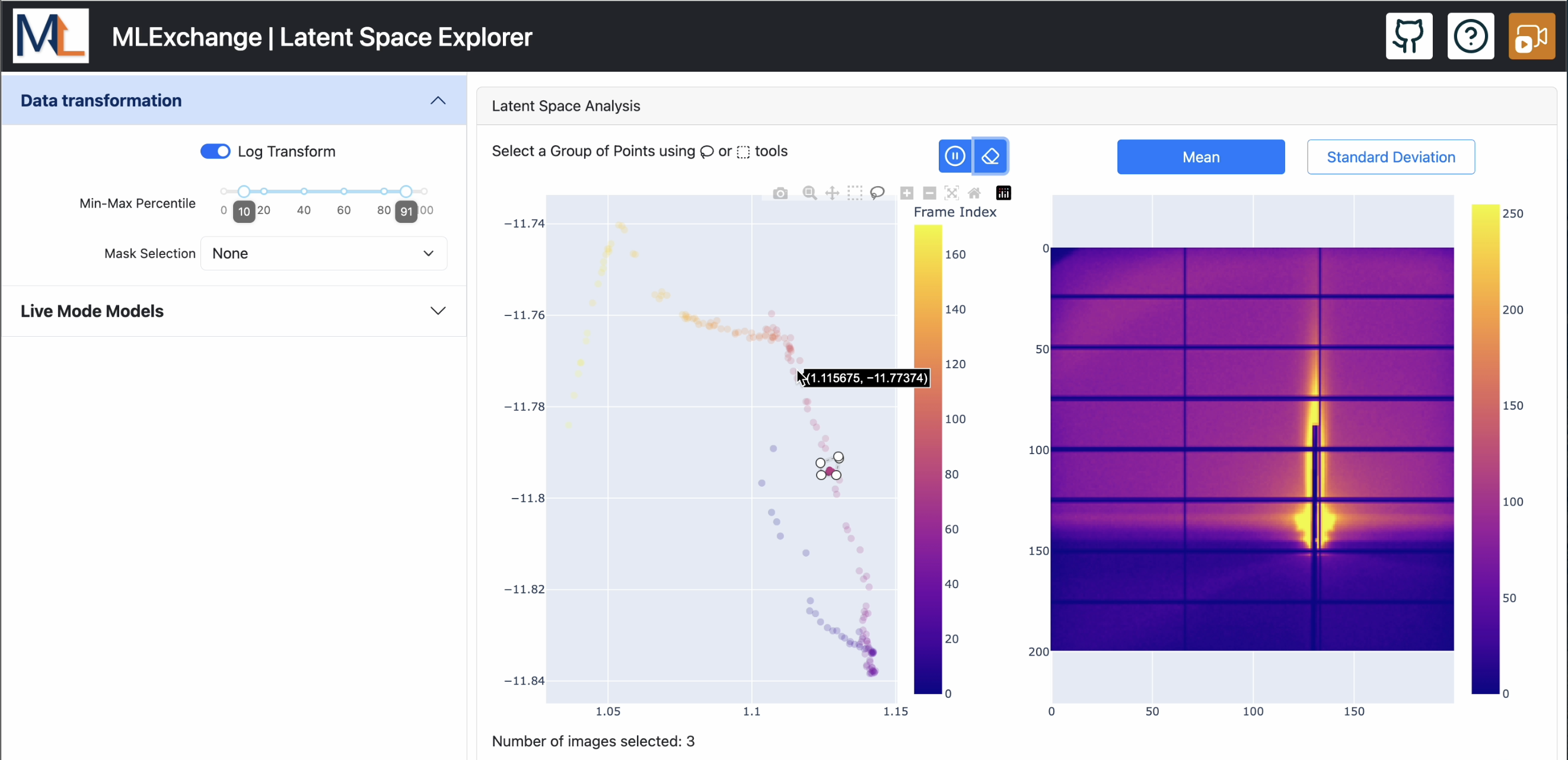}
\caption{\lsexplorer\ interface for interactive exploration of learned latent embeddings during
  NSLS-II experiments. A selected region of the latent space (left) yields mean or standard deviation
  of the corresponding scattering patterns (right).}
\label{fig:lse_app}
\end{figure}

While the analyses above were generated offline to allow direct comparison across multiple
dimensionality reduction methods, the same representations can be explored interactively through the
\lsexplorer\ interface during live experiments, as shown in Figure~\ref{fig:lse_app}.

Together, these results demonstrate that the trained C-VAE generalises beyond the training dataset and
provides real-time organization of previously unseen scattering data at two independent synchrotron
facilities. The benchmarking against DINOv3 supports the value of domain-specific training for
on-the-fly scattering analysis, and the \lsexplorer\ interface makes these representations accessible
during live experiments.

\FloatBarrier

\subsection{Synthetic Scattering Image Generation via Latent Manifold Sampling}
\label{subsec:data_generation}

\begin{figure}[!htbp]
  \centering
  \includegraphics[width=\linewidth, frame]{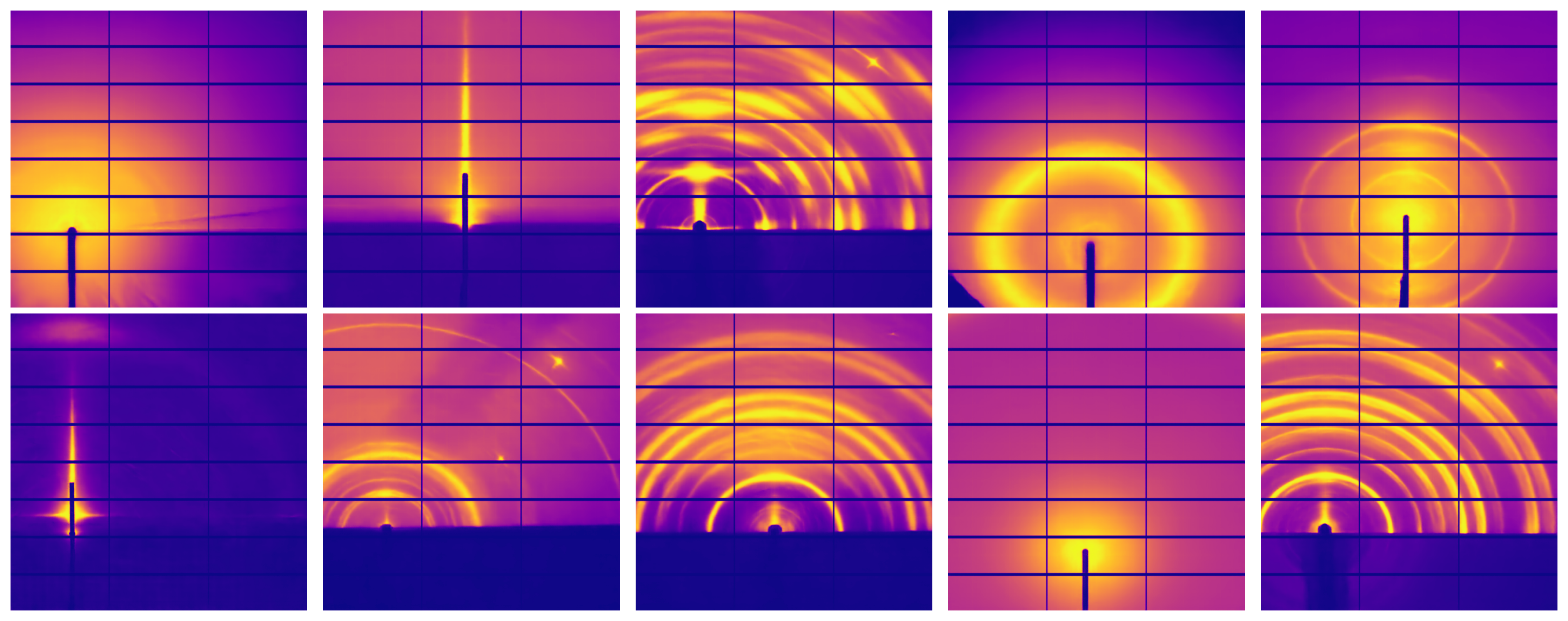}
  \caption{Synthetic scattering images generated via latent sampling of C-VAE latent space.
    Diverse structures confirm that the sampling preserves structural variety present in the training
    distribution.}
  \label{fig:gen_images}
\end{figure}

The structured latent space learned by the C-VAE supports controlled generation of synthetic scattering
images through conditioned latent sampling. Two complementary strategies are implemented: UMAP-guided
PCA sampling, which constructs cluster-aware PCA models from core latent vectors and draws new samples
using a temperature-controlled top-k softmax weighting scheme over nearby clusters in UMAP space
($k=2$, $T=0.1$), and conditional flow matching (CFM), which trains a UMAP-conditioned velocity
network to learn a continuous normalizing flow~\cite{lipman2022flow}. Classifier-free guidance~\cite{ho2022classifier}  is incorporated during training, with Ordinary Differential Equation (ODE) subsequently integrated over 20 Euler steps with guidance scale 5.0 at inference. Figure~\ref{fig:gen_images} shows synthetic scattering images generated
by both strategies across diverse structural clusters. The generated images display physically plausible
features including concentric diffraction rings, diffuse halos, and anisotropic intensity distributions
consistent with the training distribution, and occupy the same regions of the UMAP embedding as
training data, confirming that the sampling strategies preserve manifold structure without
out-of-distribution drift. A quantitative comparison of both strategies against retrieval and
unconditional baselines using a cluster-stratified evaluation protocol is provided in Supplementary
Note~2, demonstrating that flow matching achieves superior conditioning fidelity while both strategies
produce structurally realistic outputs across the full diversity of scattering patterns.

\section{Discussion}
\label{sec:discussion}

We trained a domain-specific attention-based C-VAE on 1.5 million X-ray scattering images collected at
the Advanced Light Source beamline 7.3.3, and applied it to two distinct settings: post-experiment
exploration of large experimental archives and on-the-fly analysis during live experiments. The coherent
cluster structure and smooth latent trajectories emerging from a decade of ALS measurements suggest that
dominant structural variation in X-ray scattering data is low-dimensional, a finding that justifies the
use of compact 512-dimensional embeddings for real-time analysis without significant loss of
discriminative information. Successful deployment at NSLS-II without retraining indicates that the
dominant structural variation captured by the C-VAE is facility-independent, likely reflecting the
physics of scattering rather than detector or beamline-specific artifacts. The advantage over DINOv3
(ViT-7B) in cluster separation and trajectory smoothness further supports this interpretation: natural
image features such as texture and color contrast may not be well suited for scattering-specific
structure like diffraction ring symmetry, peak sharpness, and orientation, which the domain-trained
model learns directly.

The \lsexplorer\ application within the \mlex\ platform makes these representations accessible during
both offline and live experiments. With an inference time of approximately 0.05~seconds per image, the
system supports real-time monitoring of structural transitions, inspection of representative patterns,
and interactive selection of regions of interest in the latent space as data arrive. For
higher-throughput scenarios, additional compute resources can be allocated to scale inference to faster
acquisition rates. The learned latent space supports synthetic scattering image generation through two
complementary strategies. UMAP-guided PCA sampling requires no additional training beyond the C-VAE
itself and provides broad coverage of the scattering phase space, making it practical for exploratory
applications. Conditional flow matching achieves superior conditioning fidelity, producing synthetic
images that more closely match target structural states, at the cost of a dedicated training step. Both
strategies preserve manifold structure without out-of-distribution drift, as confirmed by quantitative
evaluation across diverse structural clusters (Supplementary Note~2). Beyond data exploration, these
capabilities have practical utility for augmenting underrepresented structural states in
class-imbalanced training sets, pre-testing on-the-fly analysis pipelines before beamtime, and
generating synthetic experimental trajectories for experiment planning.

Several directions are worth pursuing in future work. The most immediate extension is integrating the
learned latent representations with autonomous experimental control. The C-VAE already detects
structural transitions in real time, and its latent trajectories could inform adaptive responses during
an experiment. These include adjusting acquisition parameters when a transition is detected, redirecting
the beam to a region of interest, or automatically initiating a complementary measurement when an
unexpected structural state appears. Moving in this direction would shift the system from passive
monitoring toward active, data-driven experimental decision making.

A second direction is incorporating experimental metadata into the learned representation. The current
model encodes only detector images, and integrating sample metadata such as temperature, humidity,
solvent composition, or deposition speed into the latent space would allow the model to distinguish
structural states that look similar in scattering but arise from different experimental conditions. This
would make the latent trajectories more physically interpretable and could help reveal correlations
between processing parameters and structural outcomes that are not visible from scattering data alone.

Third, while the ALS-trained model generalised to NSLS-II without retraining, the degree to which
facility-specific fine-tuning improves latent organization remains an open question. Transfer learning
from the pre-trained C-VAE using a small number of images from a new facility or sample system could
reduce the cost of adapting the model to new experimental contexts. Understanding how few examples are
needed to recover well-organized latent structure at a new beamline would make the approach practical
for broader deployment across the synchrotron community.

Finally, the structured outputs of the latent space analysis provide a natural interface for language
model and agent-based workflows. Current large language models struggle to reason directly over
high-dimensional detector images, but they can readily process structured descriptions of experimental
state such as cluster assignments, trajectory positions, detected transitions, and anomaly flags, which
are exactly the outputs that \lsexplorer\ already produces in real time. Connecting these
representations to a language model agent could support natural language interaction with ongoing
experiments. A scientist could ask whether the current scattering state resembles a known phase, request
a comparison between the current run and historical trajectories, or trigger alerts for specific
structural transitions. More broadly, an agent with access to both the latent space and experimental
metadata could assist with experimental planning by suggesting acquisition parameters based on the
current structural state or retrieving relevant experiments from the archive. The latent clusters and
trajectories built in this work provide the structured foundation that such a system would require.


\section*{Acknowledgements}
This work was performed and partially supported by the U.S. Department of Energy (DOE), Office of
Science, Office of Basic Energy Sciences, Data, Artificial Intelligence and Machine Learning at DOE
Scientific User Facilities program under the MLExchange Project (Award No.~107514). Support was also
provided by the Center for Materials for Water and Energy Systems (M-WET), an Energy Frontier Research
Center funded by DOE, Office of Science, Basic Energy Sciences under Award No.~DE-SC0019272. This
research used resources of the Advanced Light Source, a DOE Office of Science User Facility under
contract No.~DE-AC02-05CH11231; the National Synchrotron Light Source II, a DOE Office of Science User
Facility operated by Brookhaven National Laboratory under Contract No.~DE-SC0012704; and the National
Energy Research Scientific Computing Center (NERSC), a DOE Office of Science User Facility, under
NERSC Award No.~BES-ERCAP0027412. S.V.R. acknowledges financial support from the German
Bundesministerium f\"{u}r Bildung und Forschung (now: Bundesministerium f\"{u}r Forschung, Technologie
und Raumfahrt (BFTR)) within the ErUM-Data framework under grant No.~13D22CH7 (``Versatile Inverse
Problem Framework''). The authors thank the staff at ALS beamline 7.3.3, NSLS-II SMI beamline, and
PETRA~III P03 (MiNaXS) beamline for support during experimental data collection. GPT-5 from OpenAI and
Claude Sonnet 4.6 from Anthropic were used for minor text editing purposes in this manuscript.

\section*{Author contributions}
M.C.: Conceptualization, Data curation, Formal analysis, Investigation, Methodology, Software,
Visualization, Writing -- original draft, Writing -- review \& editing.
X.C.: Software, Writing -- review \& editing.
R.J.: Software.
W.K.: Data curation, Investigation, Software, Writing -- review \& editing.
P.H.Z.: Conceptualization, Formal analysis, Methodology, Software, Writing -- review \& editing.
D.E.: Resources, Software.
G.M.S.: Data curation, Investigation, Writing -- review \& editing.
E.S.: Investigation, Resources.
C.Z.: Investigation, Resources.
M.N.: Investigation, Resources.
N.P.W.: Data curation, Investigation, Resources.
K.K.L.: Data curation, Investigation, Writing -- review \& editing.
J.M.C.: Data curation, Investigation, Writing -- review \& editing.
J.C.D.: Data curation, Investigation, Writing -- review \& editing.
C.M.: Data curation, Investigation.
L.K.: Data curation, Investigation.
B.F.: Funding acquisition, Investigation.
G.F.: Data curation, Investigation.
Y.M.: Data curation, Investigation.
E.G.: Data curation, Investigation, Resources.
D.B.A.: Data curation, Investigation, Resources, Software.
F.S.: Resources.
B.S.: Data curation, Investigation, Resources, Writing -- review \& editing.
S.V.R.: Data curation, Investigation, Resources.
E.J.C.: Data curation, Resources, Writing -- review \& editing.
D.M.: Conceptualization, Project administration, Software, Supervision, Writing -- review \& editing.
T.C.: Conceptualization, Methodology, Project administration, Software, Supervision,
Writing -- review \& editing.
A.H.: Conceptualization, Funding acquisition, Methodology, Project administration, Resources,
Supervision, Writing -- original draft, Writing -- review \& editing.

\section*{Competing interests}
The authors declare no competing interests.

\section*{Data availability}

A subset of the X-ray scattering training dataset used in this study is publicly available via Globus (\url{https://app.globus.org/file-manager/collections/b7c0c9f3-3e22-47dc-a905-b03eaad665ac}; a Globus account may be required for access). The full training dataset and synthetic generated data are available upon reasonable request from the corresponding author.

\section*{Code availability}
The \lsexplorer\ application and associated \mlex\ platform components used in this study are openly
available at \url{https://github.com/mlexchange}. The C-VAE model training and inference scripts are
also available here.

\bibliographystyle{plain}
\bibliography{references}

\newpage

\setcounter{section}{0}
\renewcommand{\thesection}{\arabic{section}}
\renewcommand{\thesubsection}{\arabic{section}.\arabic{subsection}}
\renewcommand{\thefigure}{S\arabic{figure}}
\renewcommand{\thetable}{S\arabic{table}}
\renewcommand{\theequation}{S\arabic{equation}}
\setcounter{figure}{0}
\setcounter{table}{0}
\setcounter{equation}{0}

\begin{center}
  {\LARGE\bfseries Supplementary Information}\\[0.8em]
  {\large Unlocking Latent Dimensions: Exploring Representations of\\
  Large-Scale X-ray Scattering Data using Variational Autoencoders}\\[0.5em]
  Monika Choudhary et al.
\end{center}

\section{Latent Space Structure and Representation Validation}
\label{sec:latent_validation}

A central question in applying a learned encoder to scientific image data is whether the resulting
representation captures physically meaningful structure, or merely compresses pixel information without
semantic organization. This note addresses that question from two complementary directions: first by
examining the linear structure of the latent space through PCA, and second by comparing the C-VAE
projection against a direct pixel-based UMAP baseline.

\subsection{Principal Component Analysis of the Latent Space}
\label{appendix:pca}

\begin{figure}[!htbp]
\centering
\begin{subfigure}[t]{0.29\textwidth}
  \centering
  \includegraphics[height=4.0cm,keepaspectratio]{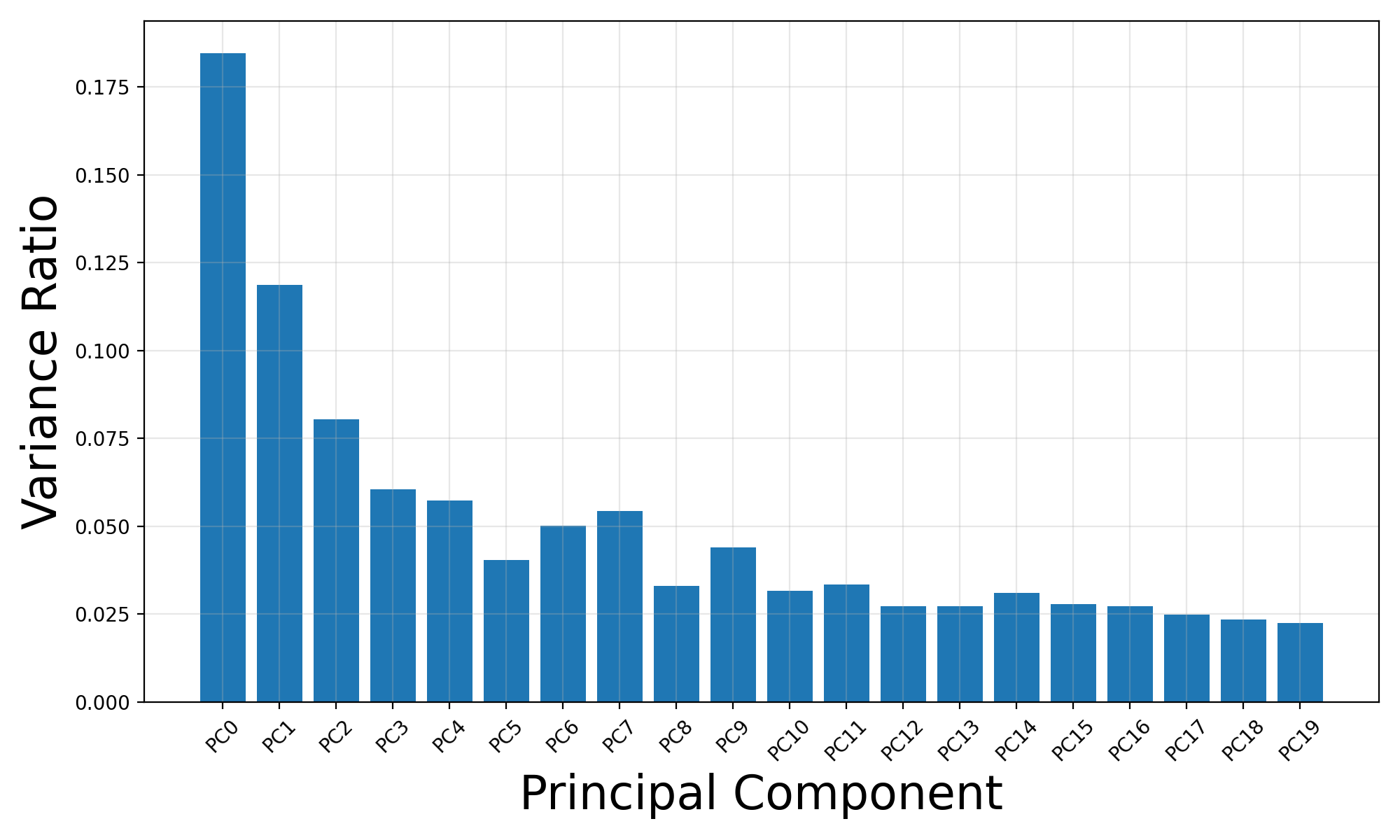}
  \caption{Variance explained by leading PCs.}
  \label{fig:variance_pc}
\end{subfigure}
\hfill
\begin{subfigure}[t]{0.29\textwidth}
  \centering
  \includegraphics[height=4.0cm,keepaspectratio]{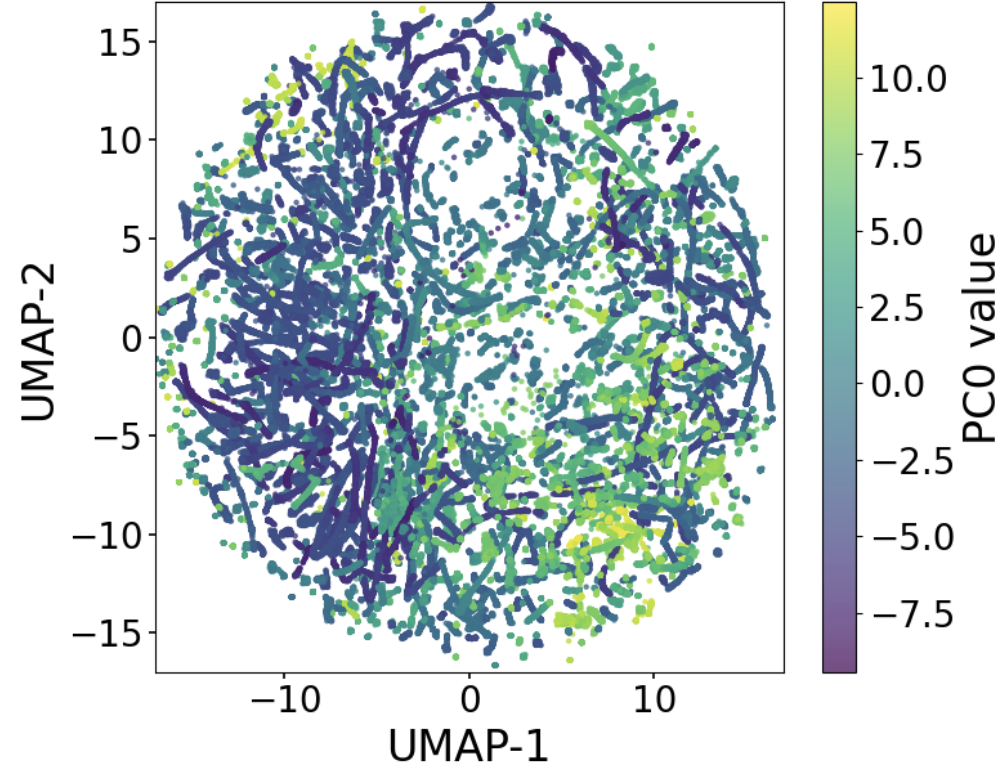}
  \caption{UMAP embedding colored by PC0 scores.}
  \label{fig:pc0}
\end{subfigure}
\hfill
\begin{subfigure}[t]{0.29\textwidth}
  \centering
  \includegraphics[height=4.0cm,keepaspectratio]{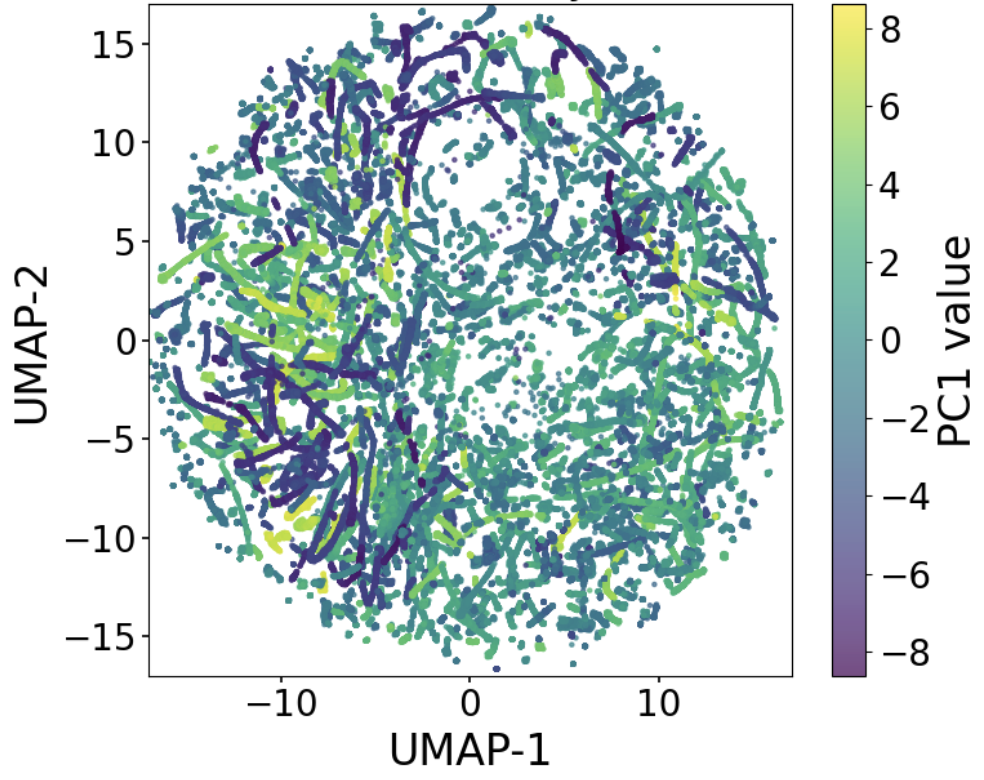}
  \caption{UMAP embedding colored by PC1 scores.}
  \label{fig:pc1}
\end{subfigure}
\caption{PCA of the C-VAE latent space from 1.5~million training images. \textbf{(a)}~Variance
  explained by the leading principal components: a small number of components capture most of the
  latent variance, confirming low-dimensional structure. \textbf{(b)}~UMAP embedding colored by PC0
  scores: the smooth gradient confirms strong alignment between the dominant linear mode and global UMAP
  structure. \textbf{(c)}~UMAP embedding colored by PC1 scores: a weaker but spatially structured
  gradient corresponding to a secondary mode of variation.}
\label{fig:pc_analysis}
\end{figure}

To characterize the linear structure of the learned representation, we perform PCA on the
512-dimensional latent vectors extracted from the 1.5~million training images.
Figure~\ref{fig:variance_pc} shows the variance explained by the leading principal components. The
first few components capture a disproportionately large fraction of the total variance, indicating that
the latent representation possesses meaningful low-dimensional structure despite its high
dimensionality. This compression is a consequence of the variational objective, which penalizes
redundant latent dimensions and encourages the encoder to represent only the dominant modes of image
variation.

To examine the relationship between linear and nonlinear structure in the latent space, the per-sample
scores along PC0 and PC1 are used to color each point in the UMAP embedding, as shown in
Figure~\ref{fig:pc_analysis}. In Figure~\ref{fig:pc0}, a clear and continuous gradient is visible
across the UMAP layout, indicating that PC0, the dominant mode of linear variation, is well-aligned
with the global structure captured by UMAP. This suggests the leading principal component corresponds
to a dominant and smoothly varying physical mode such as changes in intensity distribution, ring
curvature, or feature orientation. Figure~\ref{fig:pc1} shows a less pronounced but spatially
structured gradient for PC1, consistent with a secondary mode of variation more diffusely distributed
across the latent manifold. Together, these results confirm that the latent space organizes images along
directions that correspond to interpretable physical variation rather than arbitrary encoder outputs.

\subsection{Comparison with Direct Image UMAP}
\label{appendix:cvae_vs_image_umap}

The PCA results above confirm that the latent space is structured and low-dimensional. A complementary
question is whether this structure is necessary or whether projecting raw pixel intensities directly
through UMAP would yield equivalent groupings without a trained encoder. Direct image UMAP is appealing
in its simplicity, requiring no training or architectural choices. However, pixel similarity and
structural similarity are not equivalent in scientific imaging: two scattering patterns can share the
same morphological class while differing in brightness, beam flux, or detector noise, and the
variational objective of the C-VAE explicitly encourages the encoder to suppress such low-level
variation in favor of structural modes.

\begin{figure}[!htbp]
  \centering
  \includegraphics[width=\textwidth]{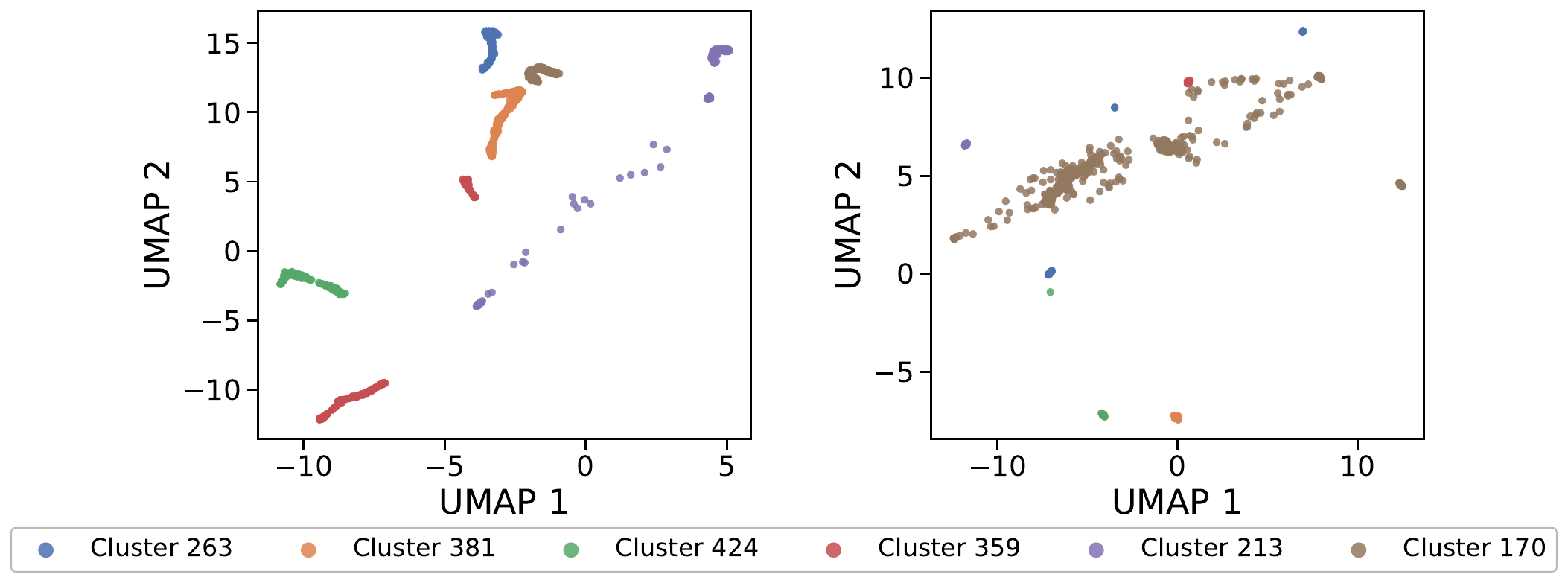}
  \caption{Comparison of UMAP projections for six representative HDBSCAN clusters (263, 381, 424, 359,
    213, 170), colored consistently across panels. Left: C-VAE latent space: each cluster forms a
    compact, well-separated manifold consistent with the structured low-dimensional organization of the
    learned representation. Right: Direct image UMAP applied to L$_2$-normalized $128 \times 128$ pixel
    vectors: the same clusters collapse into isolated points or spread without recoverable structure,
    demonstrating that raw pixel similarity is insufficient to reproduce the structural organization
    recovered by the C-VAE encoder.}
  \label{fig:cvae_vs_image_umap}
\end{figure}

To test this, each image was resized to $128 \times 128$ pixels, flattened, L$_2$-normalized to remove
global illumination differences, and projected to 2D using UMAP fitted on a random subset of
20\,000~images and applied to the full 1.5-million-image dataset. Figure~\ref{fig:cvae_vs_image_umap}
shows the result for six representative HDBSCAN clusters. In the C-VAE latent space, all six clusters
form compact, well-separated manifolds whose elongated geometry reflects the smooth temporal evolution
of scattering patterns within each run, consistent with the structured low-dimensional organization
confirmed by PCA above. In the direct image UMAP, the same clusters either collapse into isolated
points or spread without recoverable structure, even after L$_2$ normalization.

\FloatBarrier

\section{Latent Manifold Sampling Pipelines for Synthetic Scattering Image Generation}
\label{appendix:generation}

This note describes the two generation strategies evaluated in Section~2.3 of the main text:
UMAP-guided PCA sampling and conditional flow matching. Both strategies receive a two-dimensional UMAP
coordinate as conditioning input and produce a 512-dimensional latent vector that is decoded by the
trained C-VAE decoder to yield a synthetic grayscale scattering image. An overview of the UMAP-guided
PCA pipeline is shown in Figure~S3 and flow matching pipeline in Figure~S4. The conditional flow
matching architecture is summarised in Table~S1.

\subsection*{UMAP-Guided PCA Sampling}

Let $Z \in \mathbb{R}^{N \times D}$ denote the latent representation of the training data, where
$D = 512$. Each latent vector is associated with a cluster label and a cluster membership strength
obtained from the HDBSCAN clustering analysis. To restrict sampling to meaningful regions of the latent
space, only core latent vectors whose cluster membership strength meets or exceeds a threshold of 0.8
are retained. For each sufficiently populated cluster, the centroid is computed in the two-dimensional
UMAP embedding using these core samples. These centroids provide a coarse representation of the global
structure of the latent manifold.

Synthetic samples are generated by providing a query UMAP coordinate $\mathbf{u} \in \mathbb{R}^2$.
The Euclidean distance from $\mathbf{u}$ to each cluster centroid $\mathbf{c}_j$ is computed, and
these distances are converted into a probability distribution over clusters using a
temperature-controlled top-$k$ softmax weighting scheme:
\begin{equation}
    w_j = \frac{\exp\!\left(-\|\mathbf{u} - \mathbf{c}_j\|^2 \, / \, T\right)}
    {\sum_{l \in \mathcal{K}} \exp\!\left(-\|\mathbf{u} - \mathbf{c}_l\|^2 \, / \, T\right)},
    \quad j \in \mathcal{K},
\end{equation}
where $\mathcal{K}$ denotes the set of $k$ nearest cluster centroids to $\mathbf{u}$, and $T$ is the
temperature parameter. In our implementation, $k = 2$ and $T = 0.1$. Weights for all clusters outside
$\mathcal{K}$ are set to zero. A cluster is selected stochastically according to the resulting
distribution $\{w_j\}$.

To reconstruct high-dimensional latent vectors from the selected cluster, intra-cluster variability is
modelled using PCA. For each cluster, a PCA model is fitted to its core latent vectors using up to 64
space, scaled by the per-component standard deviations with a scale factor of 0.75. The sampled vector
is inverse-transformed to the full latent space to yield a candidate latent vector. To reduce
out-of-distribution artefacts, per-dimension clamping is applied, restricting each latent dimension to
the interval $\mu \pm k\sigma$, where $\mu$ and $\sigma$ denote the global mean and standard deviation
of the core latent vectors and $k = 2.5$. The resulting latent vectors are decoded in batches by the
trained C-VAE decoder to produce grayscale scattering images.

\begin{figure}[!htbp]
  \centering
  \includegraphics[width=\textwidth]{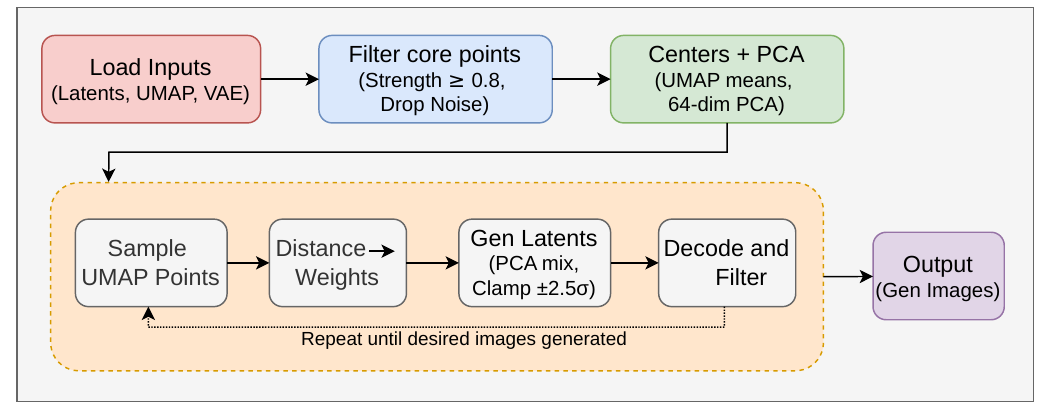}
  \caption{Overview of the UMAP-guided latent manifold sampling pipeline. Cluster-aware PCA models are
    fitted to core latent vectors, new samples are drawn using a temperature-controlled top-$k$ softmax
    weighting scheme in UMAP space, and the resulting latent vectors are decoded by the trained C-VAE
    decoder to produce synthetic scattering images.}
  \label{fig:gen_approach}
\end{figure}

\paragraph{Comparison with simpler baselines}
To contextualise this approach, three simpler baseline strategies were evaluated: global PCA sampling,
clusterwise PCA sampling, and interpolation between cluster centroids. Global PCA sampling tended to
oversmooth distinct structural modes by averaging across all clusters. Clusterwise PCA sampling limited
diversity by restricting samples to individual clusters without interpolation. Mean-based interpolation
between centroids reduced variability and produced less realistic reconstructions. Relative to these
baselines, the UMAP-guided strategy better preserved the structural diversity of the training
distribution.

\subsection*{Conditional Flow Matching}

Conditional flow matching learns a continuous normalizing flow~\cite{lipman2022flow} that transports
samples from a standard Gaussian prior $\mathcal{N}(\mathbf{0}, \mathbf{I})$ to the training latent
distribution, conditioned on a two-dimensional UMAP coordinate. This provides a more expressive
generative model than PCA-based sampling by learning the full conditional density $p(\mathbf{z} \mid
\mathbf{u})$ rather than approximating it with a Gaussian in PCA space.

\paragraph{Training objective}
Given a training latent vector $\mathbf{z}_1$ and noise $\boldsymbol{\varepsilon} \sim
\mathcal{N}(\mathbf{0}, \mathbf{I})$, the linear interpolation path is defined as:
\begin{equation}
    \mathbf{z}_t = (1 - t)\,\boldsymbol{\varepsilon} + t\,\mathbf{z}_1,
    \qquad t \sim \mathrm{Uniform}[0, 1],
\end{equation}
with velocity target $\mathbf{v} = \mathbf{z}_1 - \boldsymbol{\varepsilon}$ along this path. A
velocity network $\mathbf{v}_\theta(\mathbf{z}_t, t, \mathbf{u})$ is trained to predict this target:
\begin{equation}
    \mathcal{L}(\theta) = \mathbb{E}_{t,\, \mathbf{z}_1,\, \boldsymbol{\varepsilon},\, \mathbf{u}}
    \left\| \mathbf{v}_\theta(\mathbf{z}_t, t, \mathbf{u})
    - (\mathbf{z}_1 - \boldsymbol{\varepsilon}) \right\|^2.
\end{equation}

\paragraph{Velocity network architecture}
The network consists of six residual blocks, each applying adaptive layer normalisation (AdaLN)
conditioned on a fused 256-dimensional embedding of time and UMAP coordinate. The time coordinate is
encoded via a sinusoidal embedding of dimension 128, projected through a two-layer MLP. The UMAP
coordinate is encoded through a three-layer MLP ($2 \to 64 \to 256 \to 256$ with SiLU activations).
Both embeddings are concatenated and merged through a two-layer MLP to produce the shared conditioning
vector. Each residual block applies AdaLN normalisation, a $4\times$ expansion feedforward network with
SiLU activation and dropout of 0.1, and a residual connection. The output projection is initialised to
zero to ensure the network starts from the identity flow. Architecture hyperparameters are summarised
in Table~S1.

\paragraph{Classifier-free guidance}
To support fidelity-diversity trade-off control at inference, classifier-free
guidance~\cite{ho2022classifier} is incorporated during training by randomly zeroing the UMAP
conditioning with probability $p = 0.15$, training the network simultaneously as a conditional and
unconditional model. At inference, the guided velocity is constructed as:
\begin{equation}
    \mathbf{v}_\mathrm{guided} = \mathbf{v}_\mathrm{uncond} + s \cdot
    \left(\mathbf{v}_\mathrm{cond} - \mathbf{v}_\mathrm{uncond}\right),
\end{equation}
where $s = 5.0$ is the guidance scale. This formulation allows trading sample diversity for
conditioning fidelity at inference time without retraining.

\paragraph{Training details}
The model was trained on high-confidence cluster members with HDBSCAN membership strength exceeding
0.5, corresponding to approximately 643{,}000 points, using a 90/10 training and validation split.
UMAP coordinates were normalized to $[-1, 1]$ and latent vectors were standardized to zero mean and
unit variance prior to training. Optimization used AdamW with a learning rate of $3 \times 10^{-4}$
and weight decay of $10^{-4}$, combined with a cosine annealing learning rate scheduler and automatic
mixed precision. At inference, the ODE is integrated over 20 Euler steps with guidance scale $s = 5.0$.

\begin{figure}[h]
\centering
\includegraphics[width=\textwidth]{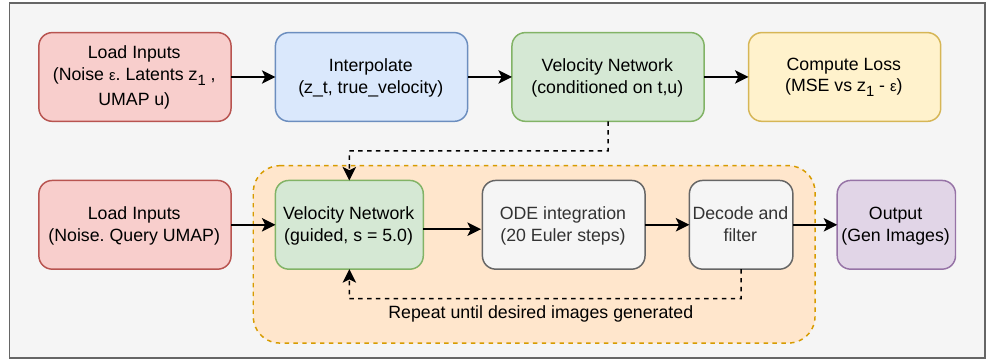}
\caption{Overview of the conditional flow matching pipeline. During training (top), the velocity
  network learns to predict the true velocity $\mathbf{z}_1 - \boldsymbol{\varepsilon}$ from
  interpolated inputs conditioned on time $t$ and UMAP coordinate $\mathbf{u}$, with classifier-free
  guidance dropout $p = 0.15$. The trained network weights are transferred directly to inference
  (bottom), where a query UMAP coordinate conditions the ODE integration over 20 Euler steps to
  produce a synthetic latent vector, which is decoded by the trained C-VAE decoder to yield a
  synthetic scattering image.}
\label{fig:cfm_pipeline}
\end{figure}

\begin{table}[h]
\caption{Conditional flow matching architecture and training hyperparameters}
\label{tab:cfm_hyperparams}
\centering
\begin{tabular}{ll}
\hline
\textbf{Hyperparameter} & \textbf{Value} \\
\hline
Latent dimension & 512 \\
UMAP conditioning dimension & 2 \\
Hidden dimension & 512 \\
Conditioning dimension & 256 \\
Number of residual blocks & 6 \\
Time embedding dimension & 128 \\
Dropout & 0.1 \\
CFG dropout probability & 0.15 \\
Guidance scale (inference) & 5.0 \\
ODE integration steps & 20 \\
Training points & ${\sim}643{,}000$ \\
Training / validation split & 90 / 10 \\
Optimiser & AdamW \\
Learning rate & $3 \times 10^{-4}$ \\
Weight decay & $10^{-4}$ \\
LR scheduler & CosineAnnealingLR \\
Training epochs & 500 \\
\hline
\end{tabular}
\end{table}

\subsection*{Quantitative Evaluation of Generation Strategies}

To evaluate generation quality, a real scattering image is selected from the training dataset at a
known UMAP location; its two-dimensional UMAP coordinate is provided to each generation method as the
sole input, and the resulting synthetic image is compared against the real image as ground truth.
Twenty real images are sampled per structural cluster, yielding a cluster-stratified evaluation set
that spans the full diversity of scattering patterns in the training distribution. We report two
complementary metrics: latent $\ell_2$ distance between the generated and real latent vectors, which
measures conditioning fidelity, and pixel mean squared error between the generated and real decoded
images, which measures image-level fidelity. Two baselines bound the comparison. PCA k-NN averages the
$k = 5$ nearest real training latents to the query UMAP coordinate and adds PCA-space noise, serving
as a retrieval upper bound with direct access to real training data. Hence, it does not produce novel
samples. PCA random samples from the global PCA distribution without any UMAP conditioning, serving as
the unconditional lower bound.

Results are summarised in Table~S2. Flow matching achieves the lowest latent $\ell_2$ distance (mean
$\pm$ s.d.: $2.48 \pm 1.82$, median 1.84), outperforming UMAP-guided PCA ($5.39 \pm 3.41$) and
approaching the retrieval upper bound set by PCA k-NN ($2.99 \pm 0.55$). On pixel MSE, PCA k-NN
scores lowest ($4.3 \times 10^{-3}$), however this reflects near-copy retrieval of real training
images rather than generative capability; flow matching ($7.6 \times 10^{-3}$) substantially
outperforms both PCA-based methods ($4.9 \times 10^{-2}$) while producing genuinely novel samples.
The unconditional baseline performs worst on both metrics, confirming that UMAP conditioning is
necessary for generating structurally relevant scattering patterns.

\begin{table}[h]
\caption{Conditional generation quality across methods (mean $\pm$ s.d. across all query points)}
\label{tab:generation_metrics}
\centering
\begin{tabular}{lcc}
\hline
\textbf{Method} & \textbf{Latent $\ell_2$ $\downarrow$} &
\textbf{Pixel MSE $\downarrow$} \\
\hline
PCA k-NN  & $2.99 \pm 0.55$ & $(4.3 \pm 6.0) \times 10^{-3}$ \\
PCA random & $6.28 \pm 1.07$ & $(4.9 \pm 4.0) \times 10^{-2}$ \\
UMAP-guided PCA & $5.39 \pm 3.41$ & $(4.9 \pm 5.9) \times 10^{-2}$ \\
Conditional flow matching & $\mathbf{2.48 \pm 1.82}$ & $\mathbf{(7.6 \pm 27.2) \times 10^{-3}}$ \\
\hline
\end{tabular}
\end{table}

\FloatBarrier

\section{Hyperparameter Selection and Evaluation Metrics}
\label{sec:hyperparams}

This note provides additional details regarding the training procedure, hyperparameter selection, and
quantitative evaluation metrics for the C-VAE model.

\subsection{Hyperparameter Selection}

Hyperparameters were chosen based on reconstruction performance, KL divergence behavior during
training, and the structural quality of the resulting latent space representations. Candidate values
and final selections are summarized in Table~\ref{tab:convvae_hparams}.

\begin{table}[htbp]
\centering
\caption{Hyperparameter configurations explored for the C-VAE model.}
\label{tab:convvae_hparams}
\begin{tabular}{llc}
\toprule
\textbf{Hyperparameter} & \textbf{Candidates} & \textbf{Chosen} \\
\midrule
Latent dimension (\textit{latent\_dim})  & 128, 256, 512              & 512 \\
Encoder/decoder depth                    & 3, 4, 5                    & 5 \\
Image size                               & 256, 512                   & 512 \\
Learning rate                            & $1 \times 10^{-4}$         & $1 \times 10^{-4}$ \\
Optimizer                                & Adam                       & Adam \\
KL weight ($\beta$)                      & 0.25, 0.5, 1.0             & 0.5 \\
Training epochs                          & 125                        & 125 \\
Training batch size                      & 16, 32, 64                 & 64 \\
Validation batch size                    & 16, 32, 64                 & 64 \\
LR scheduler                             & CosineAnnealingLR          & CosineAnnealingLR \\
\bottomrule
\end{tabular}
\end{table}

\begin{figure}[htbp]
\centering
\begin{subfigure}[t]{0.48\textwidth}
  \centering
  \includegraphics[width=\linewidth]{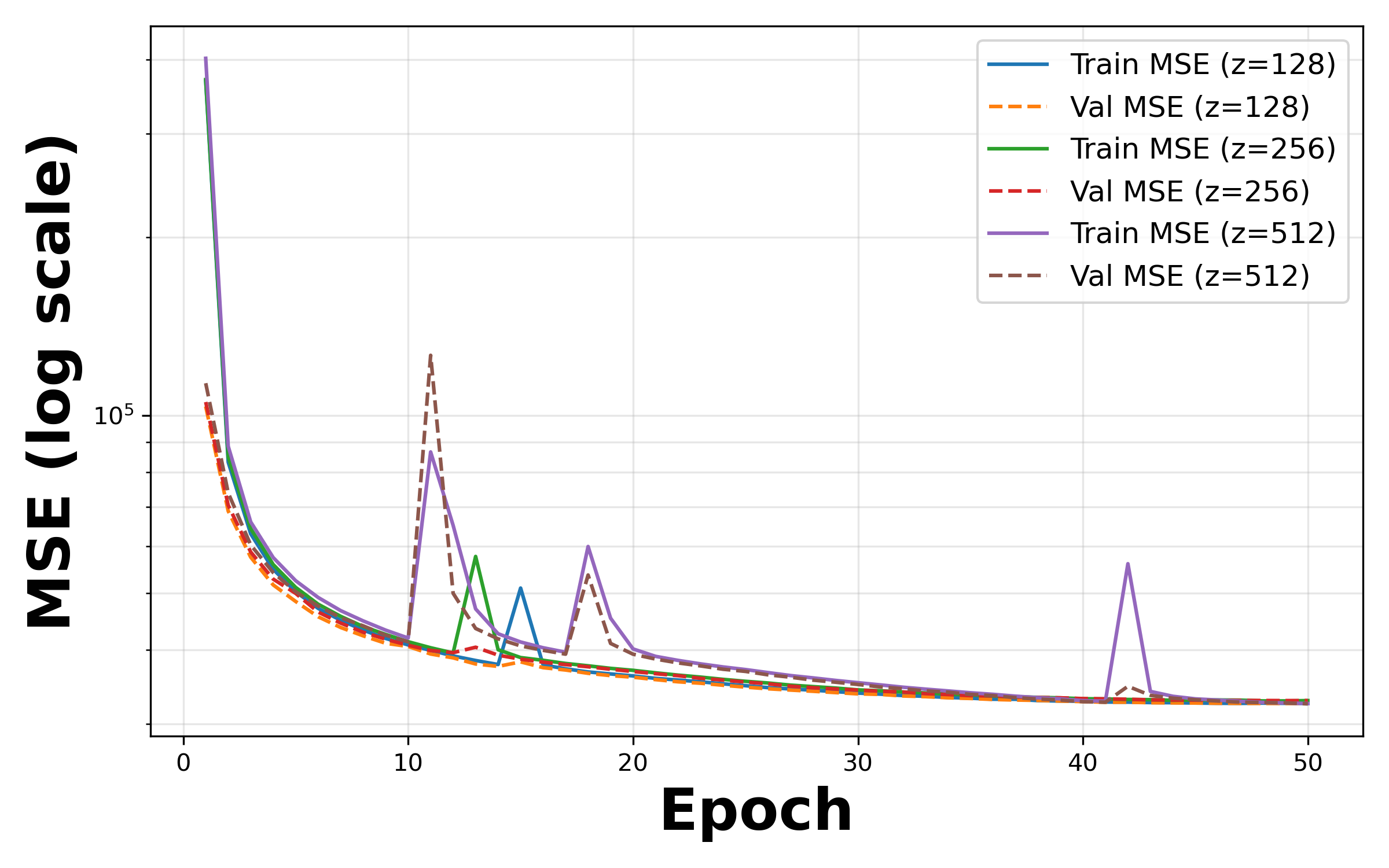}
  \caption{Reconstruction loss (MSE) plotted on a logarithmic scale for training and validation sets.}
  \label{fig:mc_recon_cvae}
\end{subfigure}
\hfill
\begin{subfigure}[t]{0.48\textwidth}
  \centering
  \includegraphics[width=\linewidth]{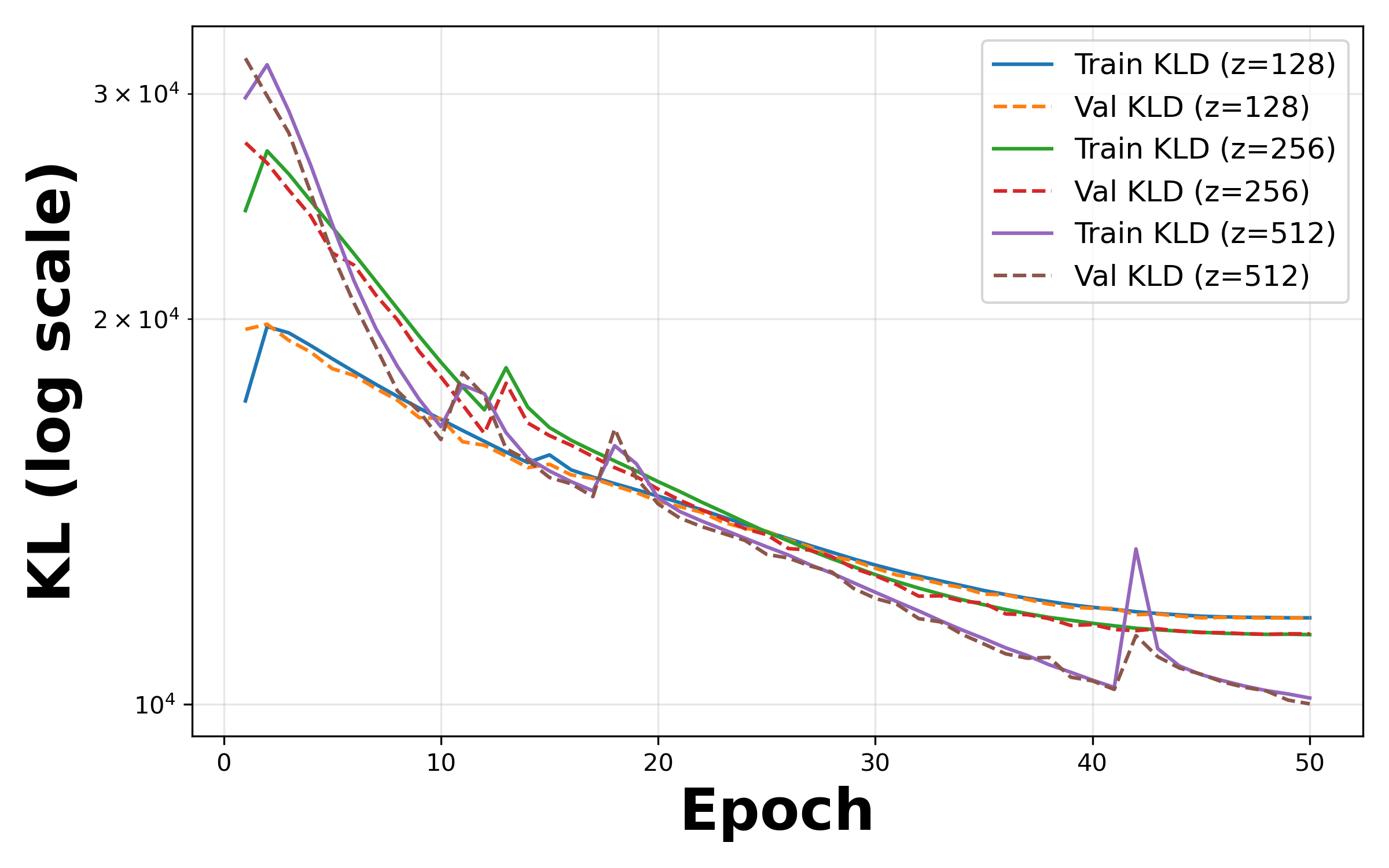}
  \caption{KL divergence loss plotted on a logarithmic scale across training epochs.}
  \label{fig:mc_kl_cvae}
\end{subfigure}
\caption{Training behavior of C-VAE models with latent dimensions of 128, 256, and 512.}
\label{fig:mcvae_training}
\end{figure}

The latent dimensionality was evaluated at 128, 256, and 512. Figure~\ref{fig:mcvae_training} shows
the reconstruction loss (MSE) and KL divergence behavior during training for all three configurations.
Increasing the latent dimension consistently reduces the reconstruction loss for both the training and
validation sets. The model with a latent dimension of 512 achieves the lowest reconstruction MSE and
demonstrates stable convergence across training epochs. Although all configurations exhibit similar
convergence trends, the 512-dimensional model maintains slightly lower and more stable KL divergence
values toward the end of training, indicating improved latent regularization.

Quantitative reconstruction and latent space metrics further support this selection, as summarized in
Table~\ref{tab:latent_metrics}. The 512-dimensional configuration achieves the lowest reconstruction
MSE and highest PSNR among the evaluated models. In addition, the latent space produced by this model
demonstrates improved clustering characteristics, indicated by a higher Silhouette score and a lower
Davies--Bouldin index. Although the latent dimensionality is set to 512, approximately 404 latent units
remain active, indicating that the model utilizes a large fraction of the available representational
capacity while maintaining effective regularization. Based on these results, a latent dimension of 512
was selected for the final model.

The encoder and decoder depth were varied between 3 and 5 layers, with a depth of 5 providing improved
reconstruction quality without signs of overfitting. All models were trained on images resized to
$512 \times 512$ pixels. During training, input images were augmented using random rotations and
horizontal or vertical flips to improve robustness. Optimization was performed using the Adam optimizer
with a learning rate of $1 \times 10^{-4}$, combined with a CosineAnnealingLR scheduler to ensure
smooth convergence throughout training. The KL divergence weight in the VAE loss function was fixed at
$\beta = 0.5$, balancing reconstruction fidelity and latent
regularization~\cite{burgess2018understanding}. Training was performed for 125 epochs using batch sizes of
64 for both training and validation datasets.

Beyond model training, the downstream dimensionality reduction steps require their own parameter
choices. Table~\ref{tab:dim_reduction_hyperparams} summarizes the settings used for PCA, UMAP, and
t-SNE applied to the learned latent representations. These are analysis decisions made after training
and do not affect the learned embeddings themselves. PCA was used as a linear baseline for variance
analysis and two-dimensional visualization. For UMAP, a higher neighbor count of 30 was selected over
smaller values to better preserve global structure across the latent space, and cosine similarity was
used as the distance metric to account for the directional nature of the L$_2$-normalized embeddings. A
minimum distance of 0.1 was chosen to allow moderate separation between clusters while avoiding
over-compression. For t-SNE, a perplexity of 30 was found to produce stable and interpretable cluster
separation. The number of iterations was set to 1500 to ensure convergence, and PCA initialization was
used to improve stability and reproducibility of the embedding. All three methods were applied to
L$_2$-normalized latent vectors extracted from the final C-VAE model.

\begin{table}[htbp]
\centering
\caption{Hyperparameter configurations for PCA, UMAP, and t-SNE.}
\label{tab:dim_reduction_hyperparams}
\begin{tabular}{llll}
\toprule
\textbf{Method} & \textbf{Hyperparameter} & \textbf{Candidates} & \textbf{Chosen} \\
\midrule
PCA   & Number of components  & --                 & 2      \\
\midrule
UMAP  & Number of neighbors   & 15, 30             & 30     \\
      & Minimum distance      & 0.05, 0.1          & 0.1    \\
      & Number of components  & --                 & 2      \\
      & Metric                & cosine, euclidean  & cosine \\
\midrule
t-SNE & Perplexity            & 20, 30, 50         & 30     \\
      & Number of iterations  & 1000, 1500         & 1500   \\
      & Initialization        & random, PCA        & PCA    \\
      & Learning rate         & --                 & auto   \\
      & Number of components  & --                 & 2      \\
\bottomrule
\end{tabular}
\end{table}

\begin{table}[ht]
\caption{Latent space and reconstruction metrics for models with different latent dimensions.}
\label{tab:latent_metrics}
\begin{adjustbox}{max width=\textwidth}
\begin{tabular}{lccccccccc}
\toprule
\makecell{Latent\\Dim} &
\makecell{Active\\Units} &
\makecell{Recon\\MSE ($10^{-3}$)} &
PSNR &
SSIM &
\makecell{KL\\Loss} &
Silhouette &
\makecell{Davies--\\Bouldin} &
\makecell{Calinski--\\Harabasz} &
\makecell{PCA\\PR} \\
\midrule
128 & 128 & 5.04 & 23.01 & 0.64 & 241.03 & 0.41 & 1.07 & \textbf{550.01} & 3.08 \\
256 & 256 & 5.18 & 22.91 & \textbf{0.65} & 237.68 & 0.40 & 1.06 & 506.91 & 3.42 \\
512 & 404 & \textbf{4.98} & \textbf{23.08} & 0.64 & \textbf{218.56} & \textbf{0.41} & \textbf{0.98} & 527.33 & \textbf{3.86} \\
\bottomrule
\end{tabular}
\end{adjustbox}
\end{table}

\subsection{Latent Space Evaluation Metrics}

To evaluate the structure of the learned latent representations, we report several commonly used
clustering and representation quality metrics.

The \textbf{Silhouette score} measures how well samples are separated between clusters. For each
sample, it compares the average distance to other points in the same cluster with the distance to
points in neighboring clusters. Higher values indicate more clearly separated clusters.

Cluster separation is further characterized by two complementary indices. The \textbf{Davies--Bouldin
index} measures the ratio of within-cluster scatter to between-cluster distance, where lower values
indicate more compact and well-separated clusters. The \textbf{Calinski--Harabasz} score measures the
ratio of between-cluster variance to within-cluster variance, where larger values indicate stronger
separation. Together they provide convergent evidence of cluster quality from different geometric
perspectives. The \textbf{PCA participation ratio} (PCA PR) estimates the effective dimensionality of
the latent representation. This metric measures how many principal components contribute significantly
to the total variance of the latent space. Higher values indicate that the latent representation
captures richer structural variability in the data.

\end{document}